\documentclass[twoside]{article}
\usepackage{aistats2014}
\usepackage{natbib}
\usepackage{todonotes}
\usepackage{listings}
\usepackage{color}
\usepackage{caption}
\usepackage{subcaption}
\usepackage{amsmath}
\usepackage{algorithm}
\usepackage[noend]{algpseudocode}

\makeatletter
\def\BState{\State\hskip-\ALG@thistlm}
\makeatother

\begin{document}

%

%

\twocolumn[

\aistatstitle{Slice Sampling for Probabilistic Programming}
\aistatsauthor{Razvan Ranca \And Zoubin Ghahramani}

\aistatsaddress{University of Cambridge \And University of Cambridge } ]

\begin{abstract}
We introduce the first, general purpose, slice sampling inference engine for probabilistic programs. This engine is released as part of StocPy, a new Turing-Complete probabilistic programming language, available as a Python library. We present a transdimensional generalisation of slice sampling which is necessary for the inference engine to work on traces with different numbers of random variables. We show that StocPy compares favourably to other PPLs in terms of flexibility and usability, and that slice sampling can outperform previously introduced inference methods. Our experiments include a logistic regression, HMM, and Bayesian Neural Net.
\end{abstract}

\section{Introduction}
There has been a recent surge of interest in probabilistic programming, as demonstrated by the continued development of new languages (eg: \cite{wood2014new, goodman2008church, lunn2009bugs, milch2007blog}) and by the recent DARPA\footnote{Probabilistic Programming for Advancing Machine Learning: {http://ppaml.galois.com/wiki/}} program encouraging further research in this direction. The increase in activity is justified by the promise of probabilistic programming, namely that of disentangling model specification from inference. This abstraction would open up probabilistic modelling to a much larger audience, including domain experts, while also making model design cheaper, faster and clearer.

Two of the major challenges lying ahead for Probabilistic Programming Languages (PPLs), before their full promise can be achieved, are those of usability and inference performance (\cite{gordon2013an}). In this paper we address usability by presenting a new PPL, StocPy, available online as a Python library (\cite{ranca2014stocpy}). We also take a step towards better inference performance by implementing a novel, slice sampling, inference engine in StocPy.

StocPy is based on the PPL design presented by \cite{wingate2011lightweight}, but is written purely in Python, and works on model definitions written in the same language. This enables us to take full advantage of Python's excellent prototyping abilities and fast development cycles, and thus allows us to specify models in very flexible ways. For instance models containing complex control flow and elements such as stochastic (mutual) recursion are easily representable. Additionally, the pure Python implementation means StocPy itself provides a good base for further experiments into PPLs, such as defining novel naming techniques for stochastic variables, looking at program recompilation to improve inference performance, or testing out new inference engines. We illustrate the benefits StocPy offers by discussing some of the language's features and contrasting the definitions of several models in StocPy against those in Anglican (\cite{wood2014new}).

We believe that ease of prototyping and implementing novel inference engines is crucial for the future success of PPLs. As decades of compiler design have shown , a ``magic bullet'' inference technique is unlikely to be found (eg: \cite[p.~401]{appel1998modern}).  This is re-inforced by the fact that PPLs can allow for a great deal of flexibility regarding the types of models and queries posed (eg: propositional logic can be represented by exposing delta functions, probabilistic programs can be conditioned on arbitrary boolean expressions). Such flexibility means a general PPL inference engine has a daunting task ahead, which includes subtasks such as handling boolean satisfiability problems. Therefore, it seems vital to develop a toolbox of inference techniques which work well on different types of modelling tasks. This high-level reasoning supports not only the development of StocPy itself, but also of the slice sampling inference engine, which outperforms previous inference techniques on certain classes of models.

\begin{figure}
  \centering
\begin{lstlisting}[language=Python, basicstyle=\ttfamily\tiny, label=alg:gausMeanModel]
import stocPy

def guessMean():
  mean = stocPy.normal(0, 1, obs=True)
  stocPy.normal(mean, 1, cond=5)

samples = stocPy.getSamples(guessMean, 10000, alg="slice")
stocPy.plotSamples(samples, xlabel="Mean")
\end{lstlisting}
  \caption{Complete StocPy program that infers a gaussian's mean.}
  \label{fig:gaussMeanModel:1}
\end{figure}

\section{StocPy}
We implement a novel PPL, StocPy, which is made available online (\cite{ranca2014stocpy}). StocPy has both a Metropolis-Hastings inference engine (implemented in the style presented by \cite{wingate2011lightweight}) and a Slice sampling inference engine which we further discuss in Section \ref{sec:Slice}. Weighted combinations of the two inference strategies can also be used.

\subsection{Why make a Python PPL?}
One of the main promises of PPLs is to allow quick and cheap development of probabilistic models, including by domain experts who may not be very comfortable with machine learning, or even with programming. The popularity of PPLs such as BUGS is partly due to their simplicity. On the other hand, lisp-based PPLs such as \cite{goodman2008church} offer greater flexibility but at the price of usability. StocPy is able to offer the same power as the lisp-based PPLs while also allowing the user to work in Python, a language both very friendly to beginners and extremely popular for prototypying.

Additionally, using Python means we can make StocPy easily available as a library, and that we can make use of Python's flexible programming style when defining models (eg: stochastic mutually recursive functions, use of globals). A user need only provide a function entry point to their model, which can then call any other functions, make use of any globals defined in the file, and use any common\footnote{More exotic constructs, such as the use of list comprehensions in place of for loops, are not currently supported by the automatic variable naming. However they can be used, if the stochastic primitives involved are manually named by the user.} Python constructs as needed. 

Finally, using Python offers both the user and the language designer a rich library support. For instance, as language designers, we are able to provide support for all 96 stochastic primitives defined in the ``scipy.stats'' library\footnote{scipy.stats primitives: \\ http://docs.scipy.org/doc/scipy/reference/stats.html} by defining a single wrapper function. Without this library, all 96 primitives would have had to be specified, in a computationally efficient way, in the language itself, which would have required a significant amount of effort. The library support also reduces the barrier of entry for all manner of PPL research. For instance, Python's excellent network and graph libraries (eg: \cite{hagberg-2008-exploring}) could be used to define a novel naming convention for stochastic primitives, which takes the program's control flow into account. Such a naming convention is required by the framework of \cite{wingate2011lightweight} and could be an improvement over the simpler version presented in that paper. In general, StocPy tries to accomodate for such avenues of research (eg: StocPy's automatic variable naming can be turned off by setting a single flag).

\subsection{StocPy Programming Style}
\label{sect:progStyle}
As mentioned above, a user defined model can make use of most common Python features, as long as the model is confined to one file. The interaction with StocPy is meant to be lightweight. Essentially, the user should perform all stochastic operations via StocPy, rather than another library of random numbers. That is to say, rather than calling \texttt{scipy.stats.norm(0,1)}, the user could call either the StocPy stochastic primitive directly: \texttt{StocPy.normal(0,1)}\footnote{these are only availalble for a few distributions. See the StocPy library for details}, or use the StocPy wrapper over scipy.stats: \texttt{StocPy.stocPrim("normal", (0,1))}. By defining stochastic primitives through StocPy, the user will define a valid generative model. Conditioning is done within the same function call that defines a variable, by specifying the ``cond'' parameter (eg: \texttt{StocPy.normal(0,1,cond=True)}). Finally, the variables or expressions we'd like to observe can be specified in two ways, either directly, in the variable definitions, via the ``obs'' parameter (eg: \texttt{stocPy.normal(0,1,obs=True)}), or via a bespoke observe statement (eg: \texttt{StocPy.observe(expression, "expName")}). The later is more flexible, allowing us to observe the result of arbitrary expressions aggregated, via their name, in arbitrary ways. Once the model is defined, inference can be done by calling one of several inference functions, and passing the entry point to the model as a parameter. A minimal (but complete) model definition, which tries to infer the mean of a gaussian given a single observation, is shown in Figure \ref{fig:gaussMeanModel:1}. This model is revisited in Figures \ref{fig:3GausPost} and \ref{fig:3GausSpec} where we see its true posterior and abstract model specification respectively. 

In Figure \ref{fig:gaussMeanModel:1}, line 1 imports our library. Line 3 defines the function that is the entry point to our model. Line 4 specifies the prior on our mean (also a gaussian) and tells StocPy that we want to observe the values this variable takes. Line 5 conditions a normally distributed random variable with our sampled mean and variance 1, on our only observation (5). Line 7 performs inference on our model, extracting 10,000 samples via slice sampling. Finaly, line 8 uses a StocPy utility function to plot the resulting distribution, which is shown in Figure \ref{fig:gaussMeanModel:2}.

In Figures \ref{alg:branching}, \ref{alg:hmm} we present more examples of models expressed in StocPy and in Anglican. The models are 2 of those presented by \cite{wood2014new} (more models are included in the supplementary material). The Anglican specifications are taken either directly from the paper or from the Anglican website\footnote{Anglican model repository: \raggedright http://www.robots.ox.ac.uk/~fwood/anglican/examples/}. We can see that in the case of the more complex models, we are able to provide a more succint representation in StocPy than in Anglican.

\begin{figure*}
  \centering
  \begin{minipage}[l]{1.0\columnwidth}
\begin{lstlisting}[language=Python, basicstyle=\ttfamily\tiny, label=alg:branchPy]
def branching():
  r = stocPy.poisson(4, obs=True)
  l = 6 if r > 4 else fib(3*r) + stocPy.poisson(4)
  stocPy.poisson(l, cond=6)
\end{lstlisting}
  \end{minipage}
  \hfill{}
  \begin{minipage}[r]{1.0\columnwidth}
\begin{lstlisting}[language=Lisp, basicstyle=\ttfamily\tiny, label=alg:branchingAng]
[assume r (poisson 4)]
[assume l (if (< 4 r) 6 (+ (fib(* 3 r)) (poisson 4)))]
[observe (poisson l) 6]
[predict r]
\end{lstlisting}
  \end{minipage}
  \caption{Branching model expressed in StocPy (left) and Anglican (right)}
  \label{alg:branching}
\end{figure*}

\begin{figure*}
  \centering
  \begin{minipage}[l]{1.0\columnwidth}
\begin{lstlisting}[language=Python, basicstyle=\ttfamily\tiny, label=alg:hmmPy]
sProbs = (1.0/3, 1.0/3, 1.0/3)
tProbs = {0 : (0.1, 0.5, 0.4), 1 : (0.2, 0.2, 0.6), 2 : (0.15, 0.15, 0.7)}
eMeans = (-1,1,0)

def hmm():
  states = []
  states.append(stocPy.categorical(sProbs,obs=True))
  for ind, ob in stocPy.readCSV("hmm-data.csv"):
    states.append(stocPy.categorical(tProbs[states[ind]], obs=True))
    stocPy.normal(eMeans[states[ind]], 1, cond=ob)
\end{lstlisting}
  \end{minipage}
  \hfill{}
   \begin{minipage}[r]{1.0\columnwidth}
\begin{lstlisting}[language=Lisp, basicstyle=\ttfamily\tiny, label=alg:hmmAng]
[assume initial-state-dist (list (/ 1 3) (/ 1 3) (/ 1 3))]
[assume get-state-transition-dist (lambda (s) (cond ((= s 0) (list 0.1 0.5 0.4)) ((= s 1) (list 0.2 0.2 0.6)) ((= s 2) (list 0.15 0.15 0.7))))]
[assume transition (lambda (prev-state) (discrete (get-state-transition-dist prev-state)))]
[assume get-state (mem (lambda (index) (if (<= index 0) (discrete initial-state-dist) (transition (get-state (- index 1))))))]
[assume get-state-obs-mean (lambda (s) (cond ((= s 0) -1) ((= s 1) 1) ((= s 2) 0)))]
[observe-csv "hmm-data.csv" (normal (get-state-obs-mean (get-state $1)) 1 ) $2]
[predict (get-state 0)]
[predict (get-state 1)]
...
[predict (get-state 16)]
\end{lstlisting}
  \end{minipage}
  \caption{HMM model expressed in StocPy (left) and Anglican (right)}
  \label{alg:hmm}
\end{figure*}

\section{Slice Sampling Inference Engine}
\label{sec:Slice}
\subsection{Advantages of Slice Sampling}
Slice sampling (\cite{neal2003slice}) is a Markov Chain Monte Carlo algorithm that can extract samples from a distribution $P(x)$ given that we have a function $P^*(x)$, which we can evaluate, and which respects $P^*(x) = Z P(x)$, for some constant $Z > 0$. The idea behind Slice sampling is that, by using an auxiliary height variable $u$, we can sample uniformly from the region under the graph of the density function of $P(x)$.

In order to get an intuition of the algorithm we can consider the one dimensional case. Here we can view the algorithm as a method of transitioning from a point $(x,u)$ which lies under the curve $P^*(x)$ to another point $(x',u')$ lying under the same curve. The addition of the auxiliary variable $u$ means that we need only sample uniformly from the area under the curve $P^*(x)$.

A basic Slice sampling algorithm can be described as:

\begin{algorithm}
  \caption{Slice Sampling}
  \begin{algorithmic}[1]
    \State $ \text{Pick initial } x_1 \text{ such that } P^*(x_1) > 0 $
    \State $ \text{Pick number of samples to extract } N $
    \For {$t = 1 \to N$} 
      \State $ \text{Sample a height } u \text{ uniformly from [0, } P^*(x_t)] $
      \State $ \text{Define a segment } [x_l,x_r] \text{ at height } u$
      \State $ \text{Pick }x_{t+1} \in [x_l, x_r] \text{ such that } P^*(x_{t+1}) > u$
   \EndFor
  \end{algorithmic}
  \label{alg:sliceSamp}
\end{algorithm}

The key to Slice sampling's efficiency is the fact that the operations in lines 5 and 6 can be performed with exponential stepping out and shrinking methods. One possible implementation of these operations is shown in the supplementary material. In this way, slice sampling corrects one of the main disadvantages of MH, namely that it has a fixed step size. In MH, picking a wrong step size can significantly hamper progress either by unnecesarily slowing down the random walk's progress or by resulting in a large proportion of rejected sampled. Slice sampling, however, adjusts an inadequate step size of size S with a cost that is only logarithmic in the size of S (\cite{mackay2003information}).

While some methods to mitigate MH's fixed step size exist, such as performing pre-run tests to determine a good step-size, these adjustment are not possible in the variant of MH commonly used in PPLs. We refer to the MH proposal described in \cite{wingate2011lightweight}, which is the same as the ``RDB'' benchmark used in \cite{wood2014new}. In this MH implementation, whenever we wish to resample a variable, we run the model until we encounter the variable and then sample it from its prior conditioned on the variables we have seen so far in the current run. Therefore no fine-tuning of the proposal distribution is possible.

To get an idea of what can be gained with slice sampling over simple, single-site, MH we look at two examples, one favourable to MH and one unfavourable. In this way we can get an idea of the expected ``best'' and ``worst'' case performances of the two. We choose to look at a simple gaussian mean inference problem, and place different priors over the mean. The best case for MH is a prior that is identical to the posterior, which means MH as described above (i.e. RDB) is already sampling from the correct distribution. Conversely, the worst case for MH is an extremely uninformative prior, such that most samples will end up being rejected (in our example, the prior is uniform between 0 and 10,000 while the posterior is very close to a normal with mean 2 and variance 0.032, more details in the supplementary material).

\begin{figure*}[!ht]
        \centering
        \begin{minipage}[t]{0.32\textwidth}
           \includegraphics[width=\textwidth, height=10em]{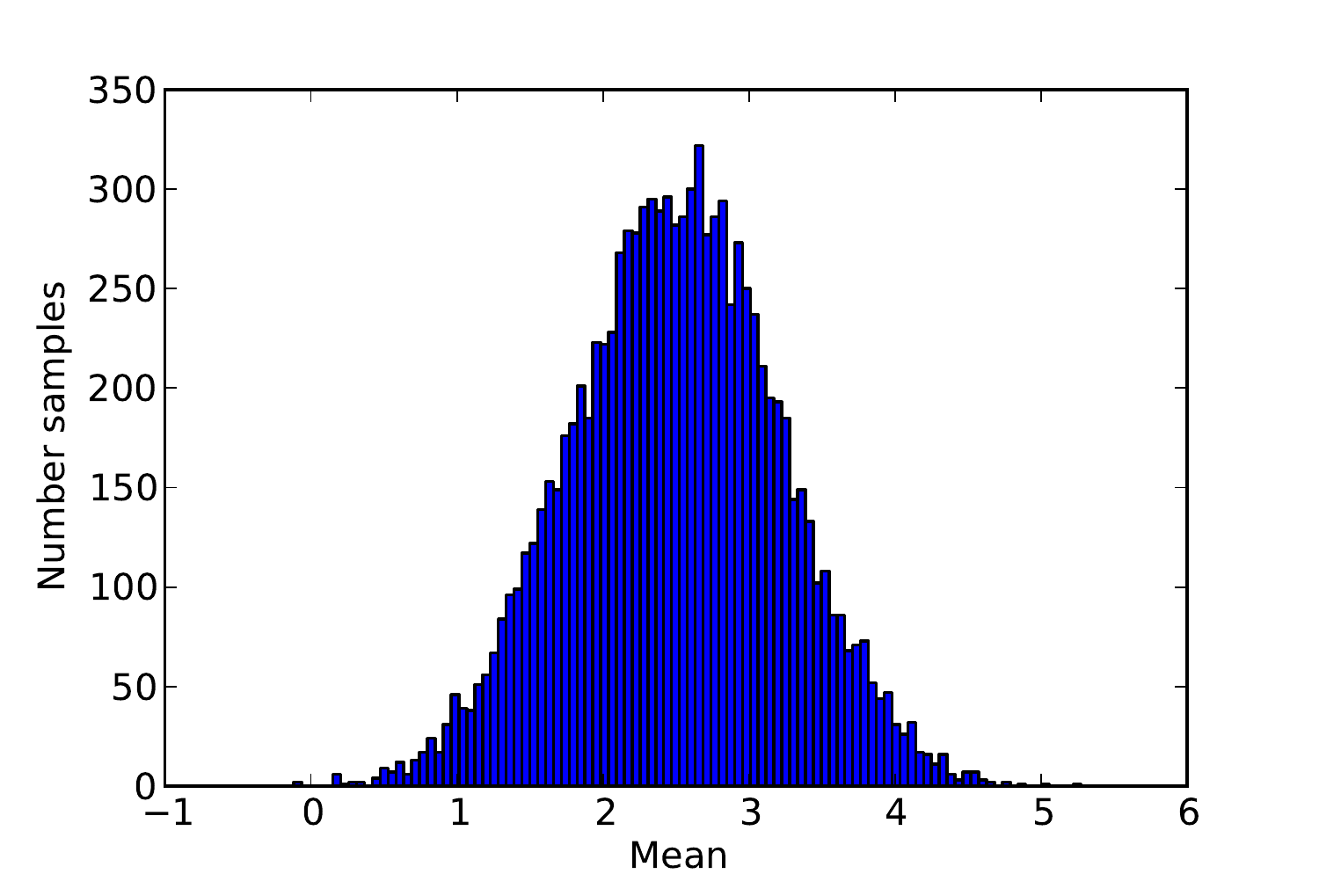}
            \caption{Inferred posterior of the model shown in Figure \ref{fig:gaussMeanModel:1}}
            \label{fig:gaussMeanModel:2}
        \end{minipage}
        ~
        \begin{minipage}[t]{0.32\textwidth}
                \centering
                \includegraphics[width=\textwidth, height=10em]{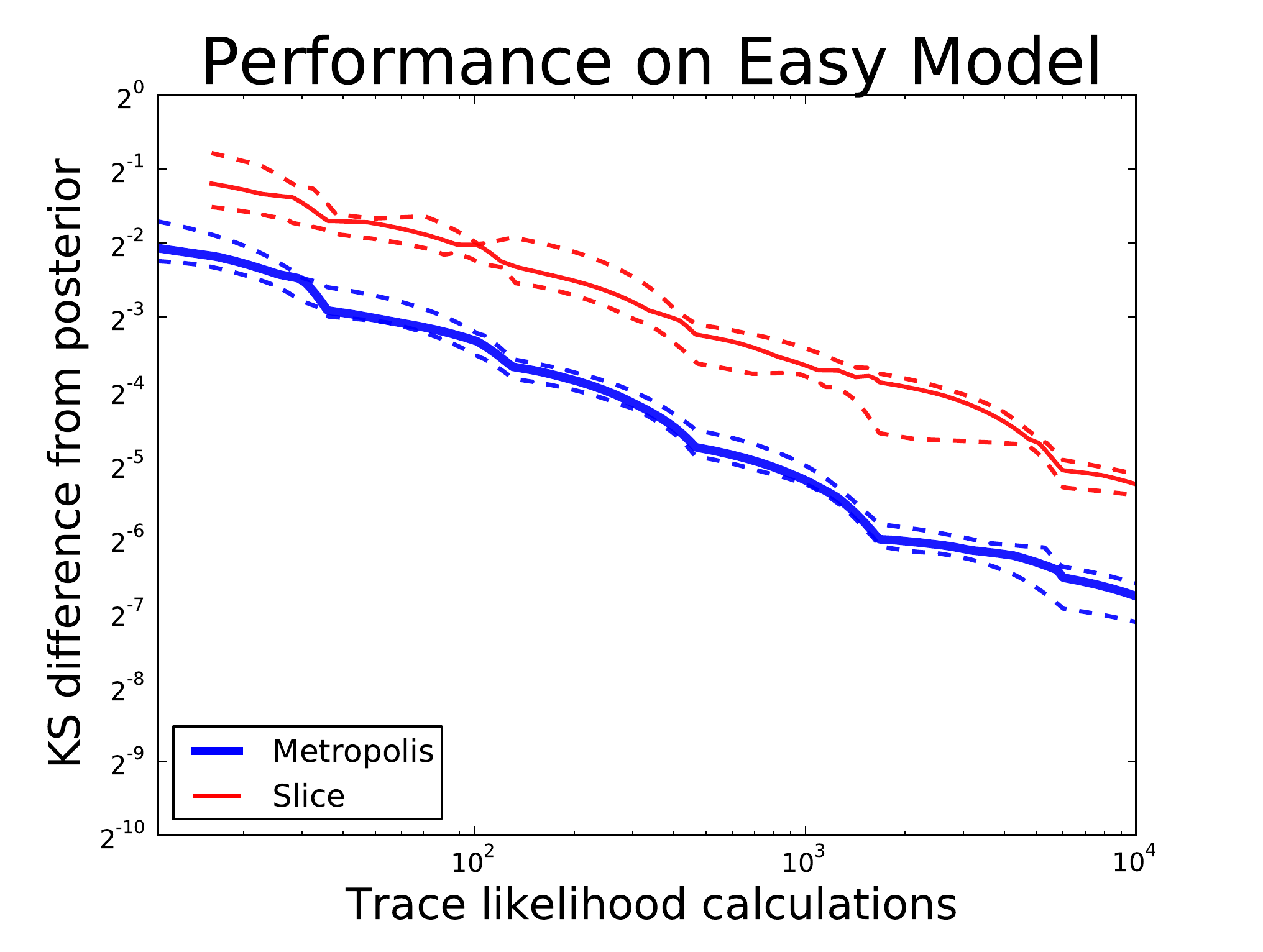}
                \caption{Case most favourable to MH, where the prior and the posterior are both equal to Normal(0, 1)}
                \label{fig:normalEasy}
        \end{minipage}
        ~
        \begin{minipage}[t]{0.32\textwidth}
                \centering
                \includegraphics[width=\textwidth, height=10em]{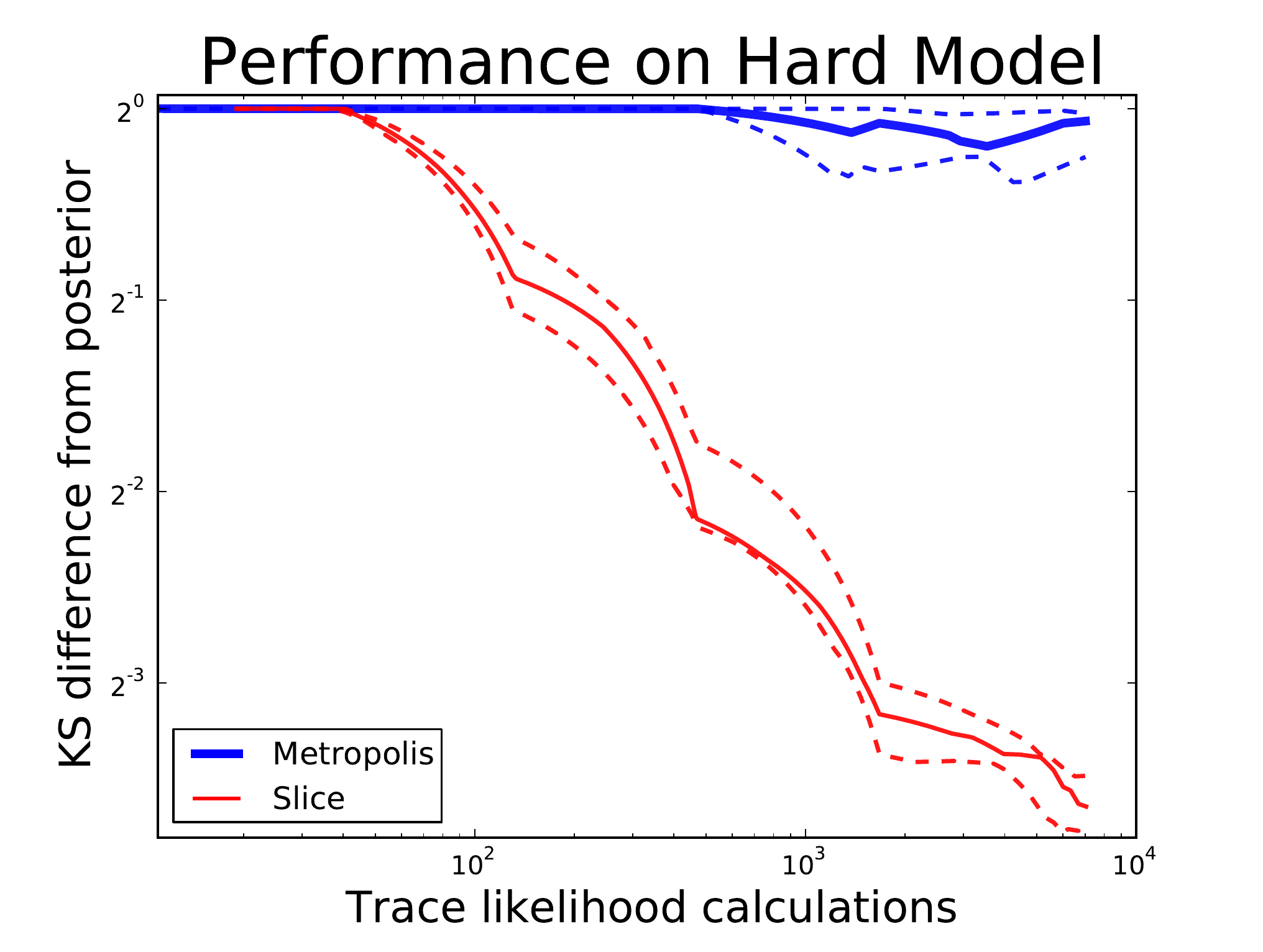}
                \caption{Case which is unfavourable to MH, where the prior is Uniform(0, 10000) and the posterior is roughly Normal(2, 0.032)
                \label{fig:normalHard}
                }

        \end{minipage}
\end{figure*}

The results are presented in Figures \ref{fig:normalEasy}, \ref{fig:normalHard}. Here we report the Kolmogorov Smirnov (KS) distance between the analytically derived posterior and a running average of the inferred posterior. The x axis shows the number of times the model has been interpreted (and had it's trace likelihood computed). We run the tests multiple times starting from different random seeds and report both the median (solid line) and the 25\% and 75\% percentiles (dashed lines) over runs. In this experiment we can see that the potential upside of slice sampling overshadows the downside. In the worst case scenario for slice sampling, we have an inference problem where our prior is equal to the posterior. That is we already know the correct answer before seeing any data. In this case the extra overhead incurred by Slice sampling slows it down roughly by a factor of 2 when compared to MH (Figure \ref{fig:normalEasy}). In the second case, however, we have a very uninformative prior and here slice sampling significantly outperforms MH (Figure \ref{fig:normalHard}). In fact, the difference between the 2 algorithms will only get more pronounced as the prior gets more uninformative. That is, examples can be created, where Slice sampling is arbitrarily faster than MH. However, we must nore that these examples are of very simple, one dimensional, unimodal models. The generalization of the insights gained from these examples to more complex models is non-trivial. Even so, we have shown a category of models where slice sampling performs substantially better than MH. Comparisons on more complex models are carried out in Section \ref{sect:empiricalEval}.

\subsection{Inference engine construction}
Our engine follows the basic slice sampling algorithm presented in Algorithm \ref{alg:sliceSamp}. As presented in \cite{wingate2011lightweight}, we consider a sample $x$ to be a program execution trace and $P^*(x)$ to be the likelihood of all stochastic variables sampled in trace $x$.

The bottleneck in the inference engines is the trace likelihood calculation. Metropolis calculates this value exactly once per sample whereas Slice sampling needs to calculate it at least 3 times (one each for $x_l$, $x_r$ and the next $x$), and potentially many more times. For this reason, StocPy provides the possibility to do inference until a certain number of trace log likelihood calculations have been performed, which allows us to compare Metropolis and slice sampling directly and fairly. Further, all experiments comparing slice with Metropolis ran the algorithms for the same length of time. We however chose to use trace likelihoods rather than seconds on the x-axis, which implicitely shows that the two engines average the same number of trace-likelihood calculation per unit of time.

The main non-trivial aspect, and novel contribution, of the inference engine construction is handling trans-dimensional models. We discuss this below.

\subsubsection{Trans-dimensional models}
\label{sect:tdModels}

In order to understand the additional complications that trans-dimensional models present for inference engines we look at a simple example taken from \cite{wood2014new}, the Branching model. This model has 2 variables whose values determine the distribution of a 3rd variable which we condition on. This model is trans-dimensional since on different traces either 1 or both of the 2 variables will be sampled.

\begin{figure*}[!ht]
        \centering
        \begin{minipage}[t]{0.48\textwidth}
                \centering
                \includegraphics[width=0.9\textwidth, height=13em]{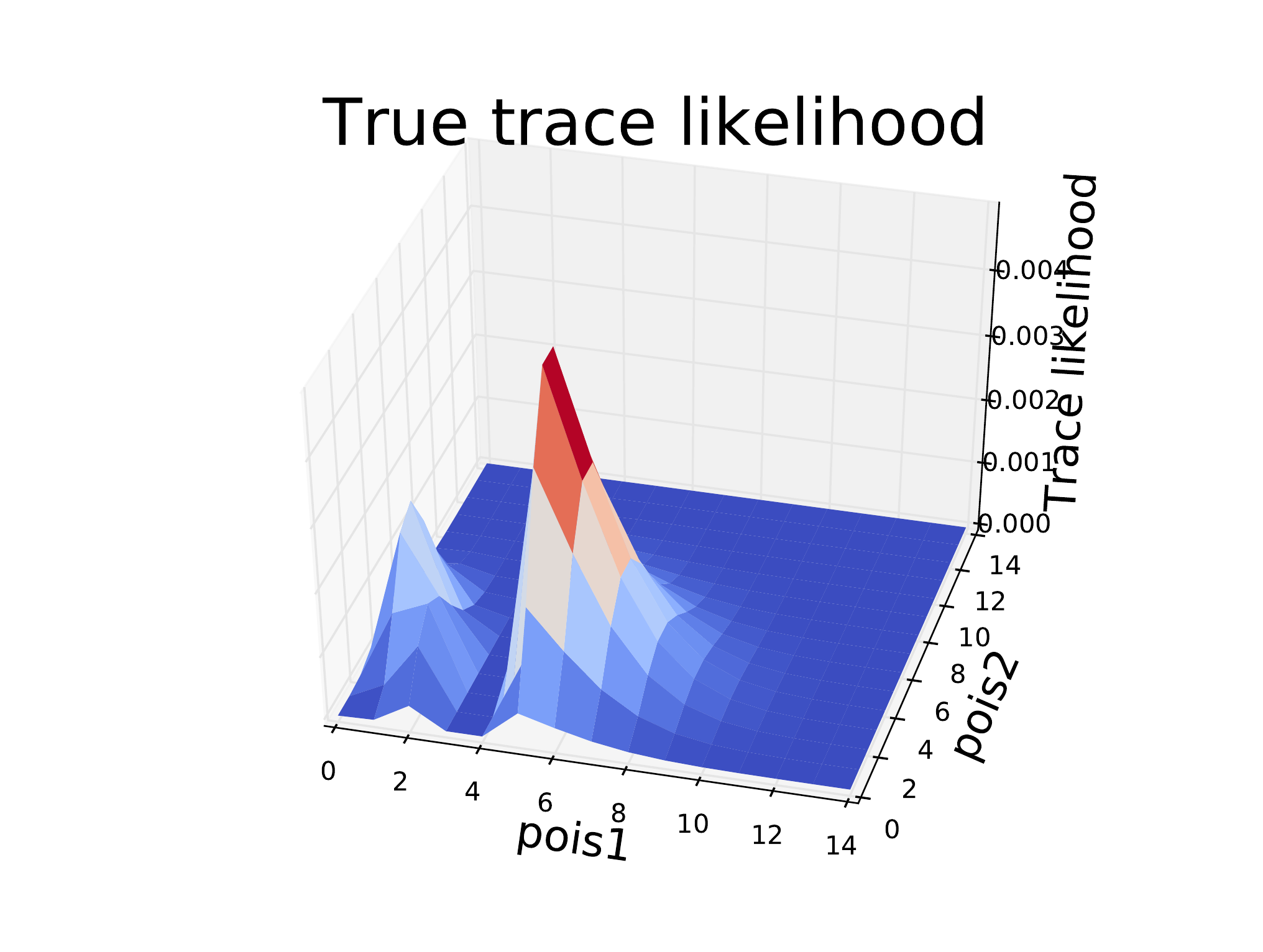}
        \end{minipage}
        ~ 
        \begin{minipage}[t]{0.48\textwidth}
                \centering
                \includegraphics[width=0.9\textwidth, height=13em]{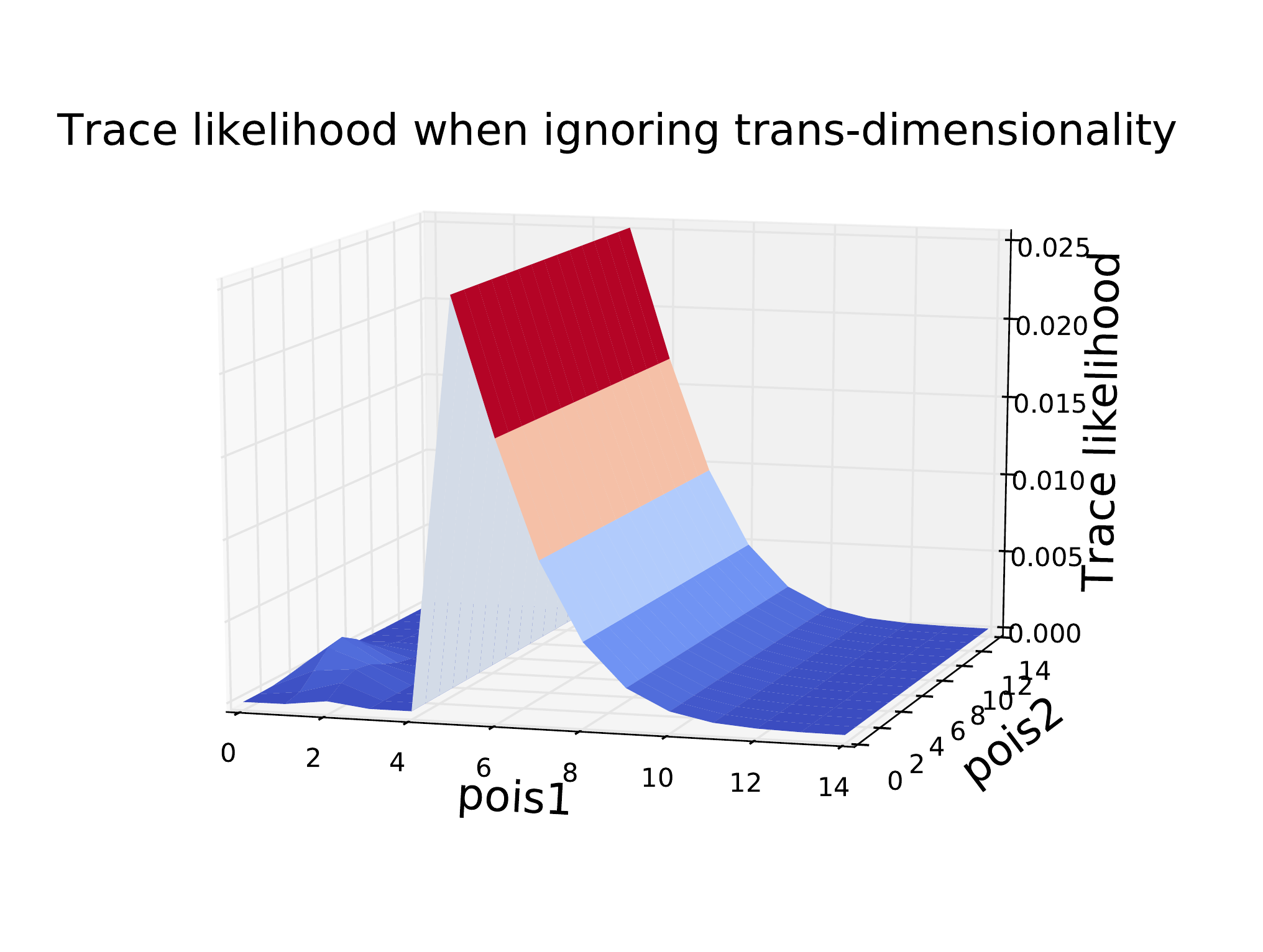}
        \end{minipage}
\caption{Likelihood space of the execution traces}
\label{fig:branchCorrTD}
\end{figure*}

Re-writing the model so that both variables are always sampled, even if one of them is unused, leaves the posterior invariant. Therefore one method to correctly perform inference in a trans-dimensional model is to always sample all variables that might be used in a trace. This approach will however be extremely inefficient in large models and is not a viable general solution. In Figure \ref{fig:branchCorrTD} we use this trick to see what the space of possible trace likelihoods looks like, and what the true posterior is. Here pois1 and pois2 refer to the Poisson variables sampled in lines 2 and 3, respectively, of the StocPy model shown in Figure \ref{alg:branching}. Integrating out the pois2 variable from the above trace likelihood space results in the correct posterior distribution.

\begin{figure*}[!ht]
        \centering
        \begin{minipage}[t]{0.48\textwidth}
                \centering
                \includegraphics[width=0.75\textwidth, height=11em]{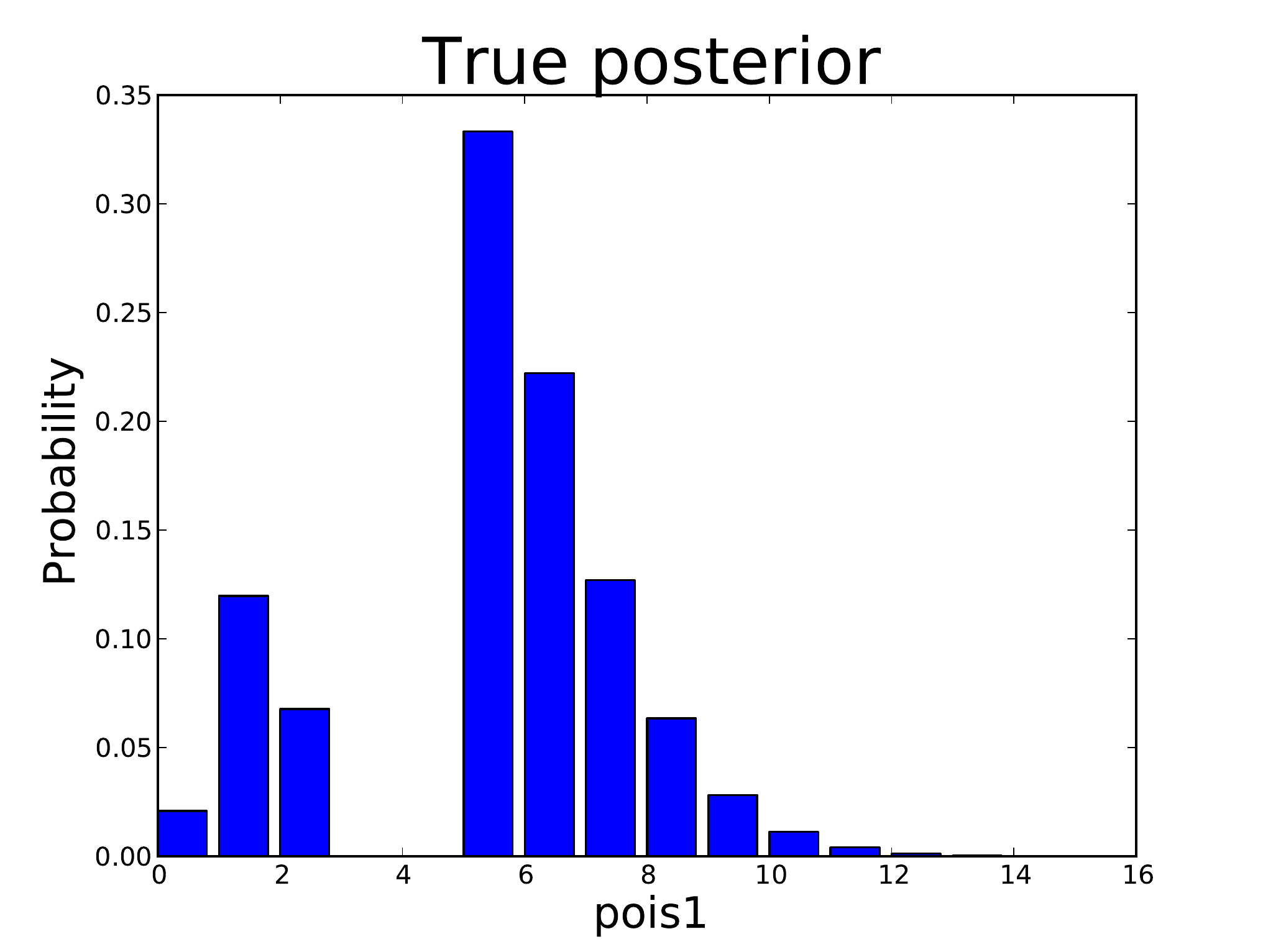}
        \end{minipage}
        ~ 
        \begin{minipage}[t]{0.48\textwidth}
                \centering
                \includegraphics[width=0.75\textwidth, height=11em]{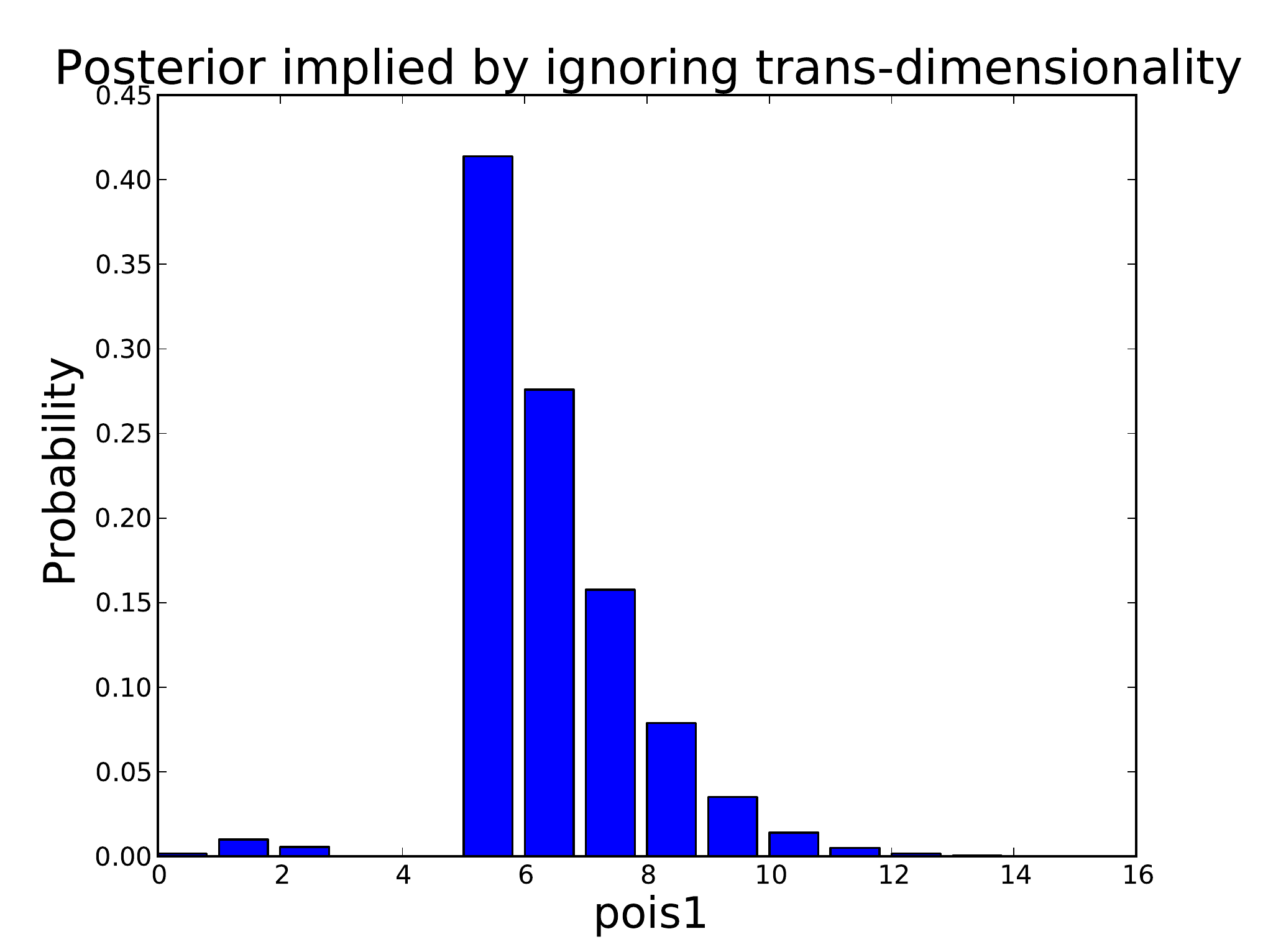}
        \end{minipage}
\caption{Posteriors inferred by integrating out pois2}
\label{fig:branchWrongTD}
\end{figure*}

The issue with trans-dimensional jumps comes from the fact that a naive inference algorithm will not sample the second poisson when it is not necessary, but will still think that the trace likelihoods of runs with different numbers of sampled variables are comparable. In doing so, the inference engine will be pretending to be sampling from a 2D trace likelihood even when it really is 1D. The space of likelihoods implied by the naive inference engine, and the posterior it would obtain by integrating out pois2, is shown in Figure \ref{fig:branchWrongTD}. We now describe how to correct the slice sampling inference engine to handle trans-dimensional models.

\subsection{Transdimensional Slice Sampling}
\label{sect:SliceTD}

To understand the trans-dimensional corrections we can place them in the framework of Reversible Jump Markov Chain Monte Carlo (RJMCMC, \cite{green2009reversible}). Informally, we can think of slice sampling as a form of RJMCMC in which we carefully choose the next sample so that the acceptance probability will always be 1. We include a short explanation of the RJMCMC notation in the Appendix.

In order to place slice sampling in this framework, we can think of a program execution trace as a state $x$. The choice of move we make is equivalent to choosing which random variable we wish to resample. Therefore, if there are $|D|$ random variables present in the current trace, and we pick one to sample uniformly, then $j_m(x) = 1/|D|$. Once the variable is chosen, we can define a deterministic function $h_m$ which produces a new value for variable $m$ by following the slice sampling specification presented in lines 4-6 of Algorithm \ref{alg:sliceSamp}. The randomness that $h_m$ must take as input is $u \sim g_m$, where $u$ is composed of $r$ random numbers and $g_m$ is the joint distribution of these $r$ numbers. The probability of moving from state $x$ to state $x'$ is then $\pi(x) g_m(u) / |D|$, and the probability of going back from $x'$ to $x$ is $\pi(x') g_m'(u') / |D'|$. Intuitively $\pi(x) = \pi(x')$ since slice sampling samples uniformly under its target distribution. Transdimensionality means that the number of variables sampled in traces will be different and therefore that $D'$ will be different from $D$ and that the dimensionality of $u$ and $u'$ ($r$ and $r'$ respectively) will also be different.

The correction we aplly to account for this is similar to the one applied by \cite{wingate2011lightweight}. Specifically we update $P^*(x) = P^*(x) + log \left( \frac{|D| * p_{stale}}{|D'| * p_{fresh}} \right)$, where $p_{stale}$ is the probability of all variables that are in the current trace but not the new and $p_{fresh}$ is the probability of variables sampled in the new trace that are not in the current. 

\section{Empirical Evaluation}
\label{sect:empiricalEval}
\subsection{Inferring the mean of a Gaussian}
\begin{figure*}[!ht]
        \centering
        \begin{subfigure}[b]{0.31\textwidth}
                \centering
                \includegraphics[width=\textwidth]{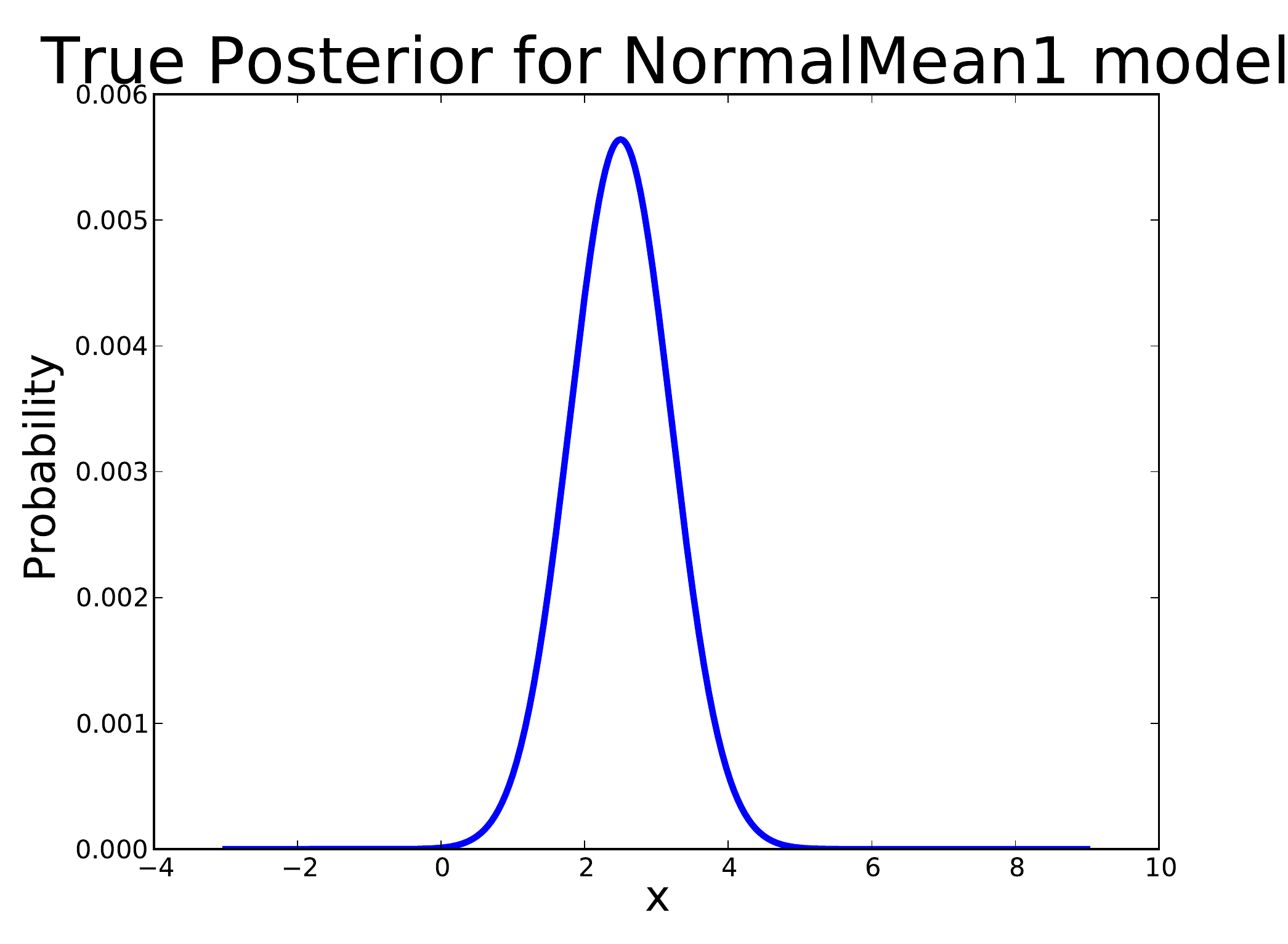}
        \end{subfigure}
        ~ 
        \begin{subfigure}[b]{0.31\textwidth}
                \centering
                \includegraphics[width=\textwidth]{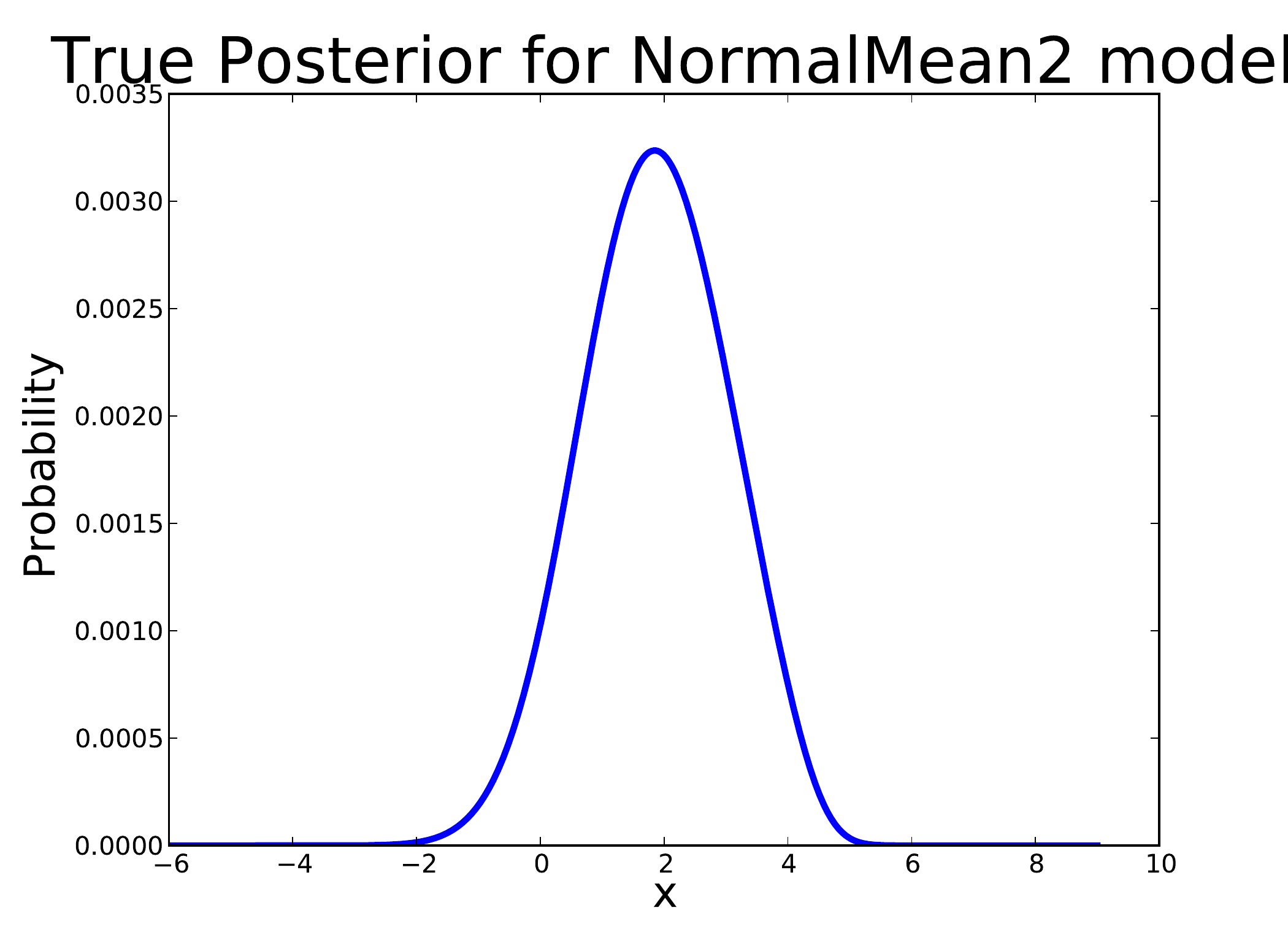}
        \end{subfigure}
        ~ 
        \begin{subfigure}[b]{0.31\textwidth}
                \centering
                \includegraphics[width=\textwidth]{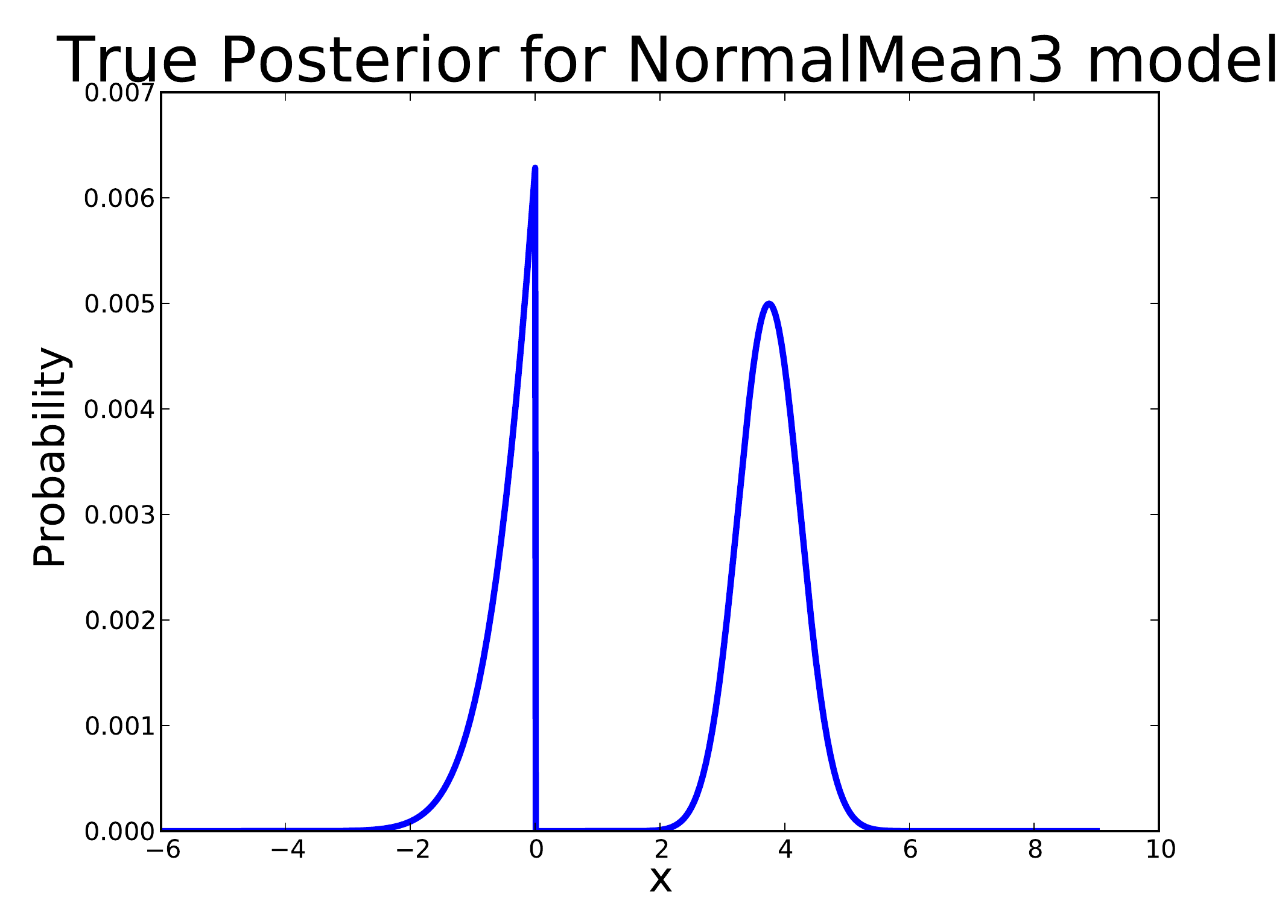}
        \end{subfigure}
    \caption{Analytically derived posteriors of the 3 gaussian mean inference models}
    \label{fig:3GausPost}
\end{figure*}

We begin with the inference of the mean of a gaussian. This time the prior and posterior are of similar sizes, but the posterior is shifted by an unusual observation. In this setting we look at 3 models, namely 1-dimensional, 2-dimensional and trans-dimensional. The model specifications are given in Figure \ref{fig:3GausSpec} and the models' posteriors are shown in Figure \ref{fig:3GausPost}.

\begin{figure*}[!ht]
  \scriptsize
\begin{minipage}[t]{.25\textwidth}
\begin{flalign*}
  &NormalMean1: &
  \\ &\quad\quad m \sim N(0,1) &
  \\ &\quad\quad \text{observe }N(m,1) = 5 &
  \\ &\quad\quad \text{predict }m &
\end{flalign*}
\end{minipage}%
\hfill{}
\begin{minipage}[t]{.25\textwidth}
\begin{flalign*}
  &NormalMean2: &
  \\ &\quad\quad m \sim N(0,1)
  \\ &\quad\quad v \sim invGamma(3,1)
  \\ &\quad\quad \text{observe }N(m,v) = 5
  \\ &\quad\quad \text{predict }m
\end{flalign*}
\end{minipage}%
\hfill{}
\begin{minipage}[t]{.25\textwidth}
\begin{flalign*}
  &NormalMean3: &
  \\ &\quad\quad m \sim N(0,1)
  \\ &\quad\quad \text{if }m < 0\text{: } v \sim invGamma(3,1)
  \\ &\quad\quad \text{else: } v = 1/3
  \\ &\quad\quad \text{observe }N(m,v) = 5
  \\ &\quad\quad \text{predict }m
\end{flalign*}
\end{minipage}
\caption{Specifications of the 3 gaussian mean inference models}
\label{fig:3GausSpec}
\end{figure*}

We now look at the performance of Metropolis, slice sampling and some different mixtures of the two over the 3 models. The mixture methods work by flipping a biased coin before extracting each sample in order to decide which inference method to use.  

\begin{figure*}[!ht]
        \centering
        \begin{subfigure}[b]{0.31\textwidth}
                \centering
                \includegraphics[width=\textwidth]{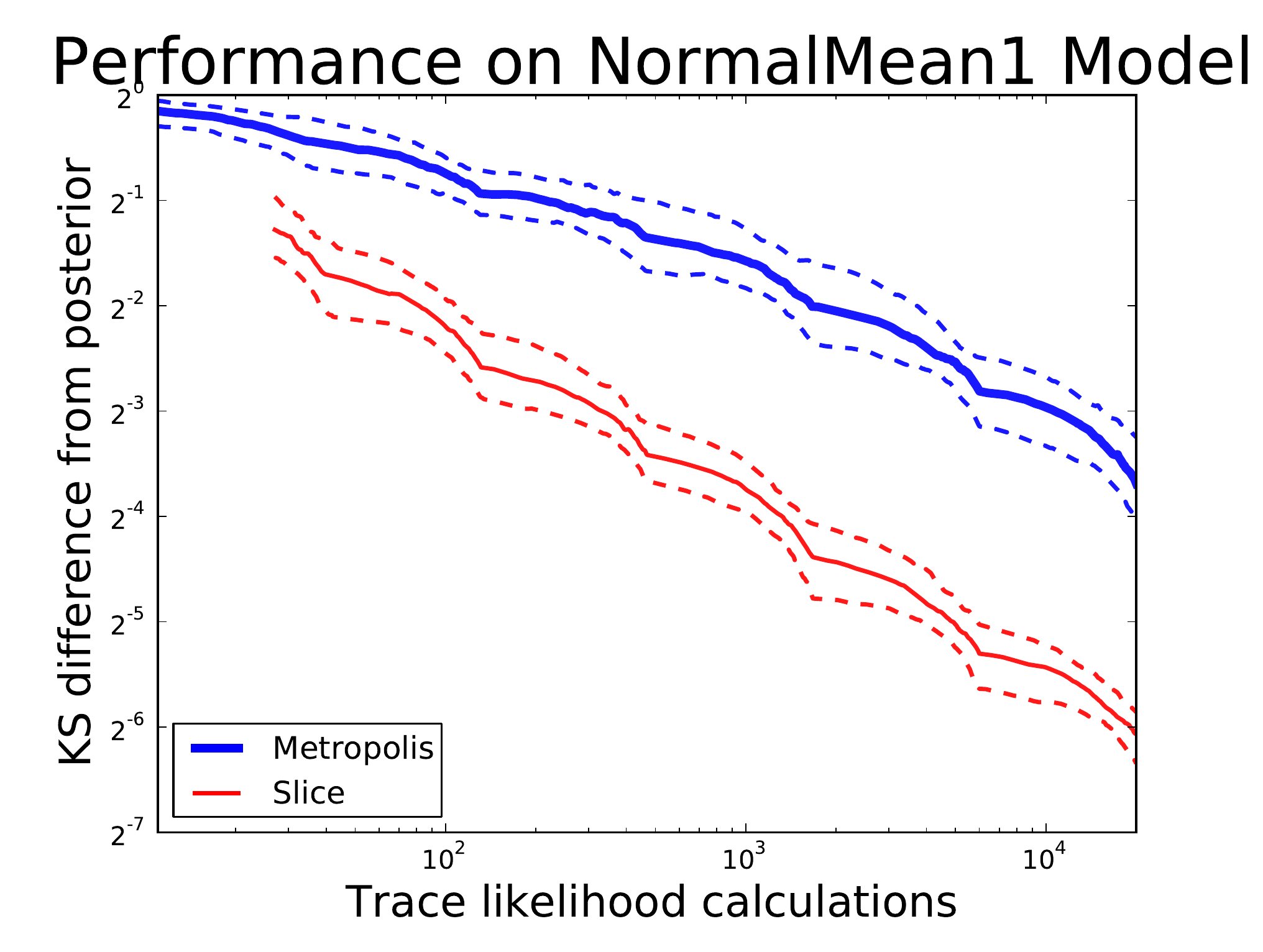}
        \end{subfigure}
        ~ 
        \begin{subfigure}[b]{0.31\textwidth}
                \centering
                \includegraphics[width=\textwidth]{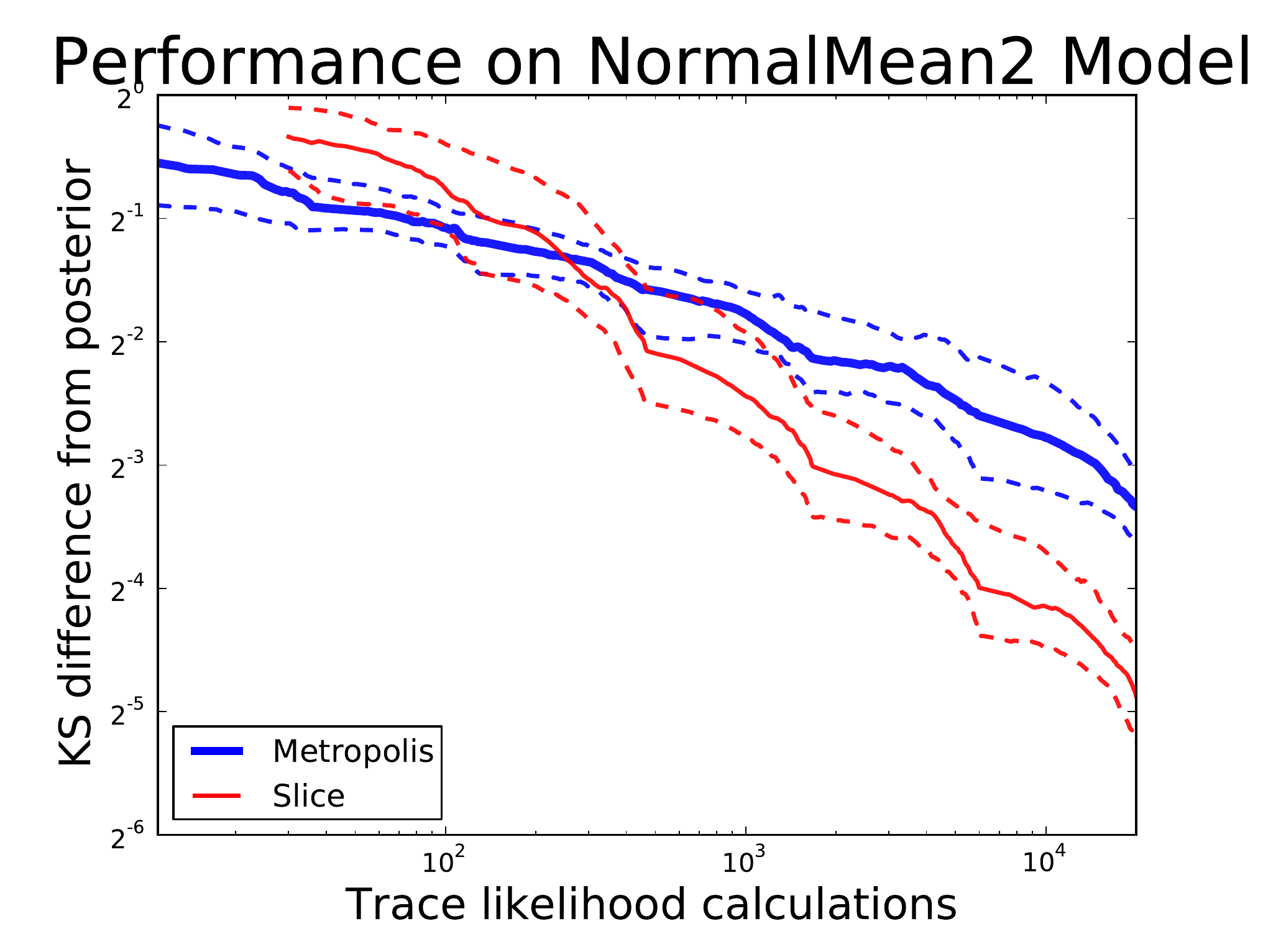}
        \end{subfigure}
        ~ 
        \begin{subfigure}[b]{0.31\textwidth}
                \centering
                \includegraphics[width=\textwidth]{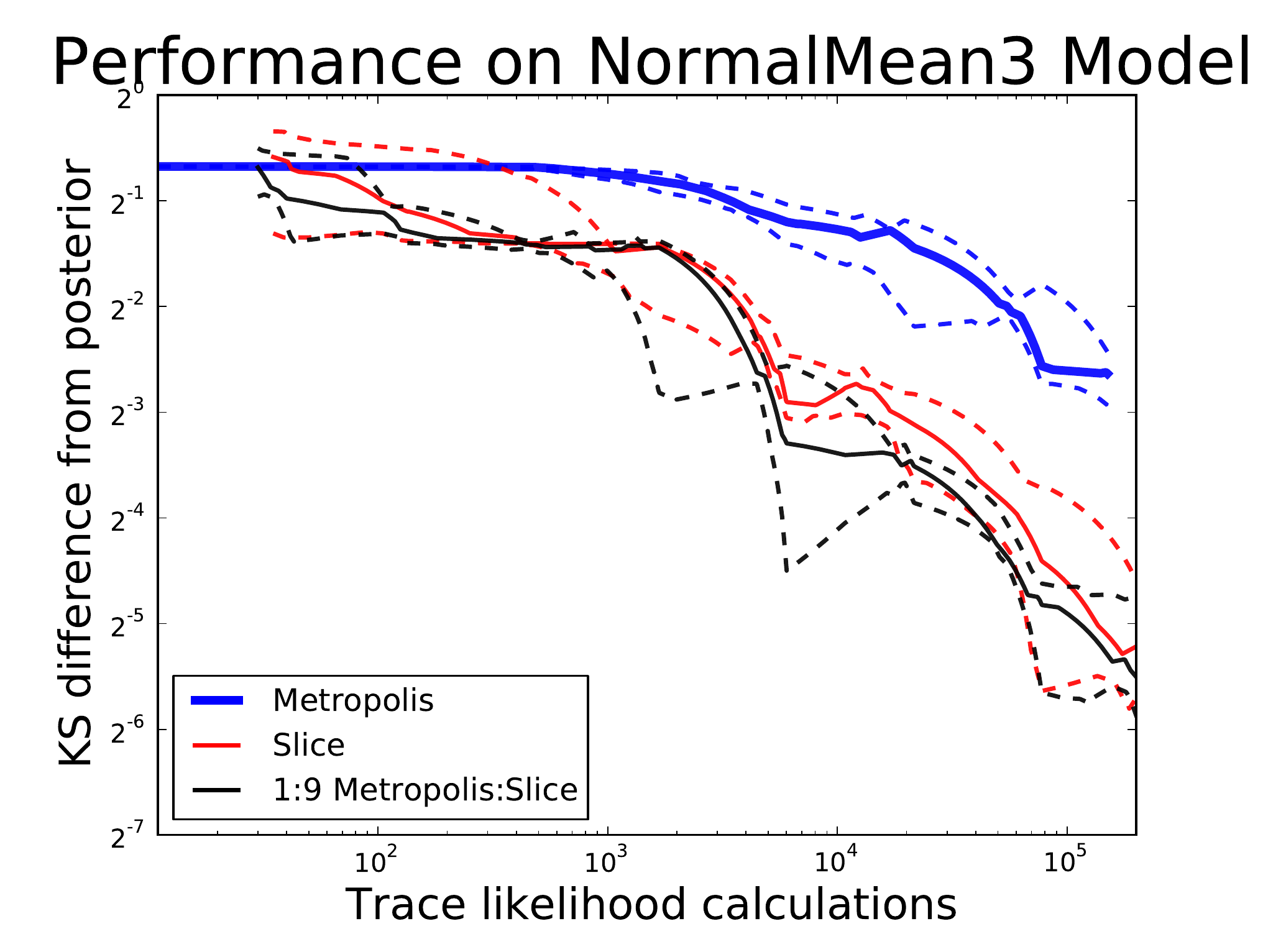}
        \end{subfigure}
    \caption{Median (solid), 25\% and 75\% quartiles (dashed) convergence rates on the 3 Gaussian models}
    \label{fig:3GausQuarts}
\end{figure*}

To compare the inference engines, we extract samples until a certain number of trace likelihood calculations are performed and then repeat this process 100 times, starting from different random seeds. We then plot the median and the quartiles of the KS differences from the true posterior, over all generated runs. Figure \ref{fig:3GausQuarts} shows the quartiles of the runs on the three models.

On the simple, 1d, model all variants of Slice sampling clearly outperform Metropolis. In the quartile graph we consider mixtures of Metropolis and Slice both with 10\% Metropolis and with 50\% Metropolis and find that the change doesn't have a significant impact on performance. This is likely because, if slice picks a good sample, Metropolis is likely to simply keep it unchanged (since it will tend to reject the proposal from the prior if it is worse).

On the 2d model, slice still clearly outperforms Metropolis, though the gap is not as pronounced as for the 1d model. Further, as in the 1d model, the 3 different slice variants all get quite similar performance. Additionally, on this model, the fact that the slice mixtures get more samples per trace calculation translates into a slightly better performance for them than for the pure slice sampling method.

The third, trans-dimensional, model reveals a more pronounced performance difference between slice and Metropolis than in the 2 dimensional case. Further, on this model, we see a significant gap between the 1:9 Metropolis:Slice method and the other slice sampling approaches. It seems that, in this case, the ratio of 1:9 strikes a good balance between slice sampling's automatic adjustment of the kernel's width and Metropolis' efficiency in likelihood calculations per sample. 

\subsection{Anglican models}
\label{sect:anglican}
In order to further test the Slice sampling inference engine we look at 3 of the models defined in \cite{wood2014new}. The model specifications for these are provided in Section \ref{sect:progStyle} and in the supplementary material. 

\begin{figure*}[!ht]
        \centering
        \begin{subfigure}[b]{0.31\textwidth}
                \centering
                \includegraphics[width=\textwidth]{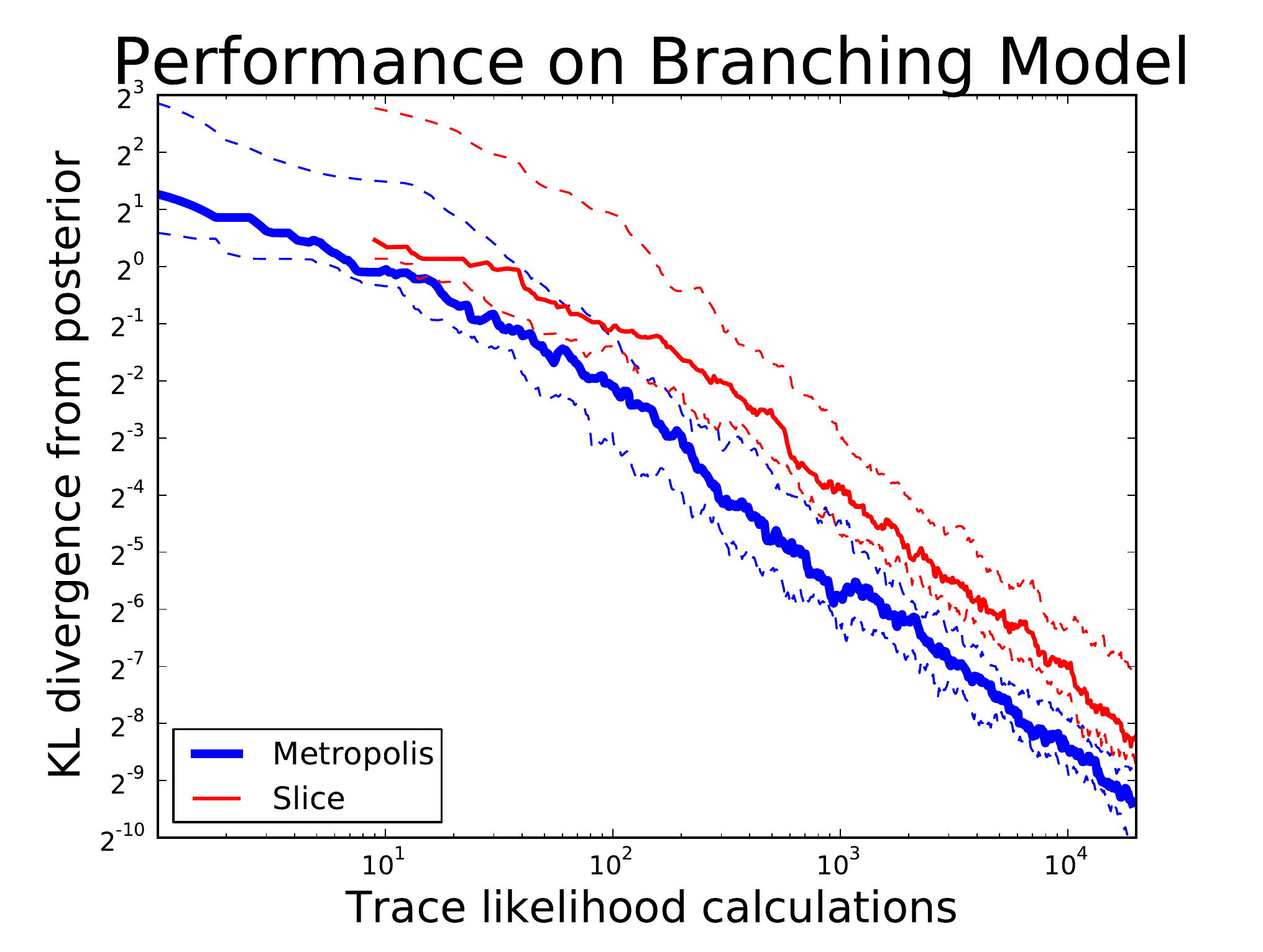}
        \end{subfigure}
        ~ 
        \begin{subfigure}[b]{0.31\textwidth}
                \centering
                \includegraphics[width=\textwidth]{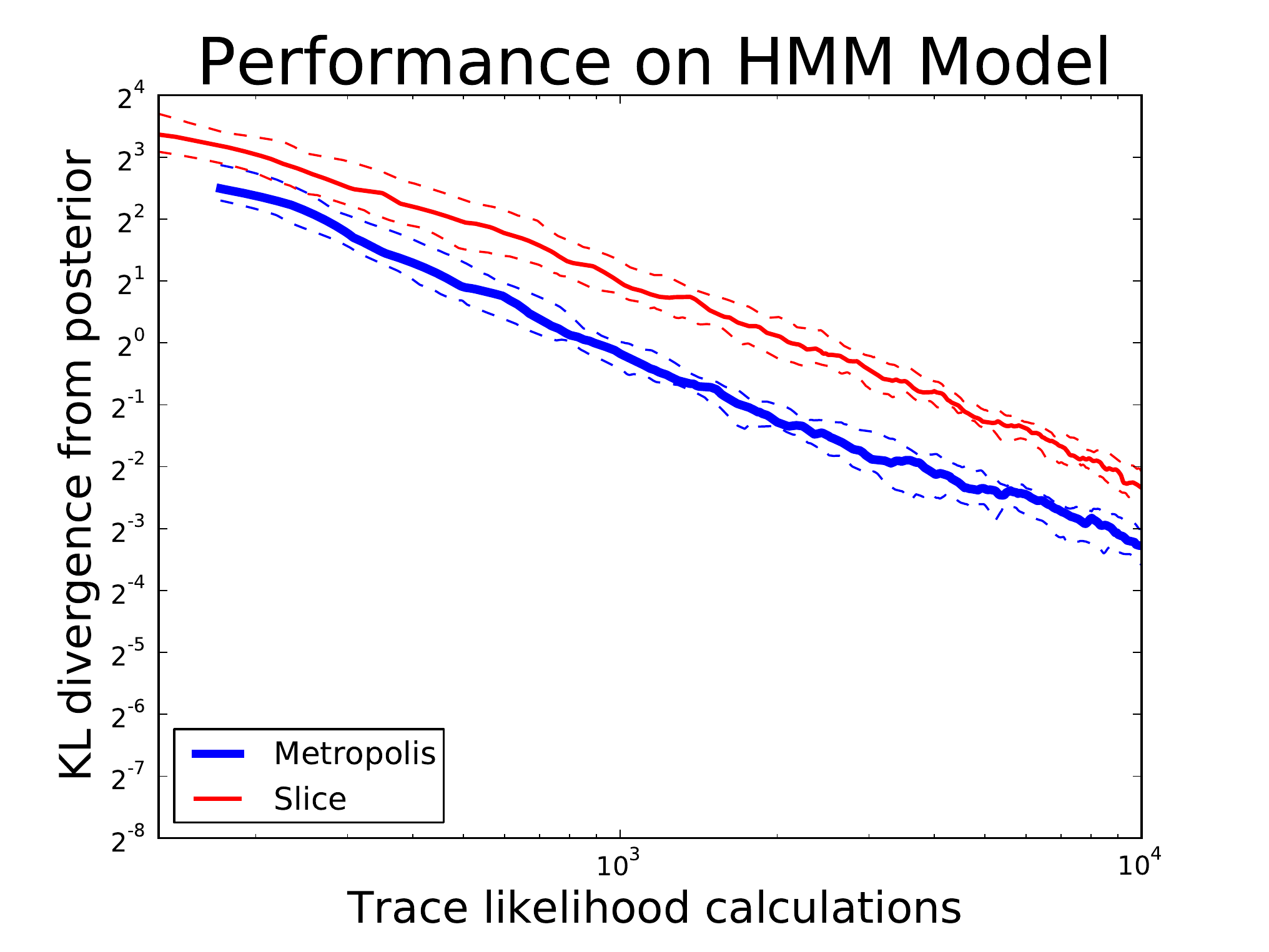}
        \end{subfigure}
        ~ 
        \begin{subfigure}[b]{0.31\textwidth}
                \centering
                \includegraphics[width=\textwidth]{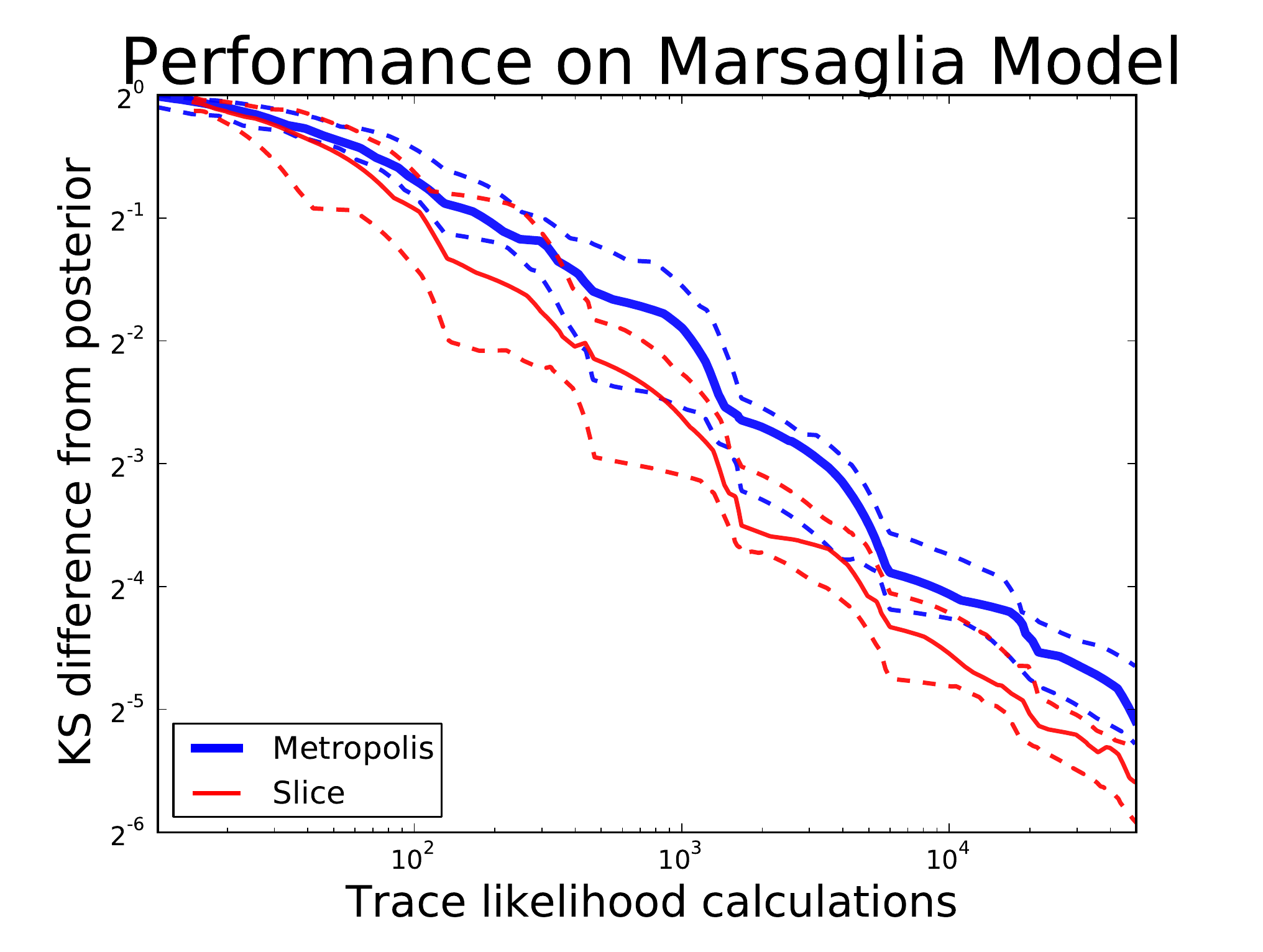}
        \end{subfigure}
    \caption{Median (solid), 25\% and 75\% quartiles (dashed) convergence rates on the 3 Anglican models}
    \label{fig:3AngQuarts}
\end{figure*}

To evaluate the engines, we use each to generate 100 independent sample runs. In Figure \ref{fig:3AngQuarts} we plot the convergence of the empirical posterior to the true posterior as the number of trace likelihood calculations increase. For continuous distributions we use the Kolmogorov-Smirnov statistic (as before), whereas for discrete ones we employ the Kullback-Leibler divergence to measure the similarity of two distributions.

On the Branching and HMM models (\ref{fig:3AngQuarts}) naive Metropolis-Hastings outperforms all Slice combinations. These models are quite small and are conditioned on few datapoints, which means they have relatively similar priors and posteriors. On such models, the overhead of slice sampling cannot be recovered, since the naive Metropolis actually does quite well by simply sampling the prior. Additionally, in the HMM model, we are only inferring in which of 3 states the model is in. With such a small state space the value that an adjusting proposal kernel can add is limited.

On the Marsaglia model however (\ref{fig:3AngQuarts}), slice sampling does obtain a boost in performance. This is due both to the continuous nature of the model (i.e. larger state space) and to the fact that the prior and posterior are significantly different (figure in Appendix). It's worth noting that the Marsaglia model is the only one in which Metropolis from the prior outperformed Anglican's Particle MCMC. Slice sampling is therefore better than both PMCMC and Metropolis on this model.

It is interesting to note that, on all models tried, slice outperforms Metropolis on a per-sample basis. This is an unfair comparison since slice does more ``work'' to generate a sample than Metropolis, but it illustrates the point that the samples chosen by slice are in general better, the only question is if the difference is sufficient as to justify the increased overhead. 

\subsection{Models for classification}
Lastly we look at some new models, namely logistic regression and a small neural network. Since we are doing classification, here we can actually plot the mean squared error that the model achieves after a certain number of trace likelihood computations. As before, we perform separate runs starting with different random seeds and plot both the median and the 25\% and 75\% quartiles. 

\begin{figure*}[!ht]
        \centering
        \begin{minipage}[t]{0.32\textwidth}
                \centering
                \includegraphics[width=\textwidth]{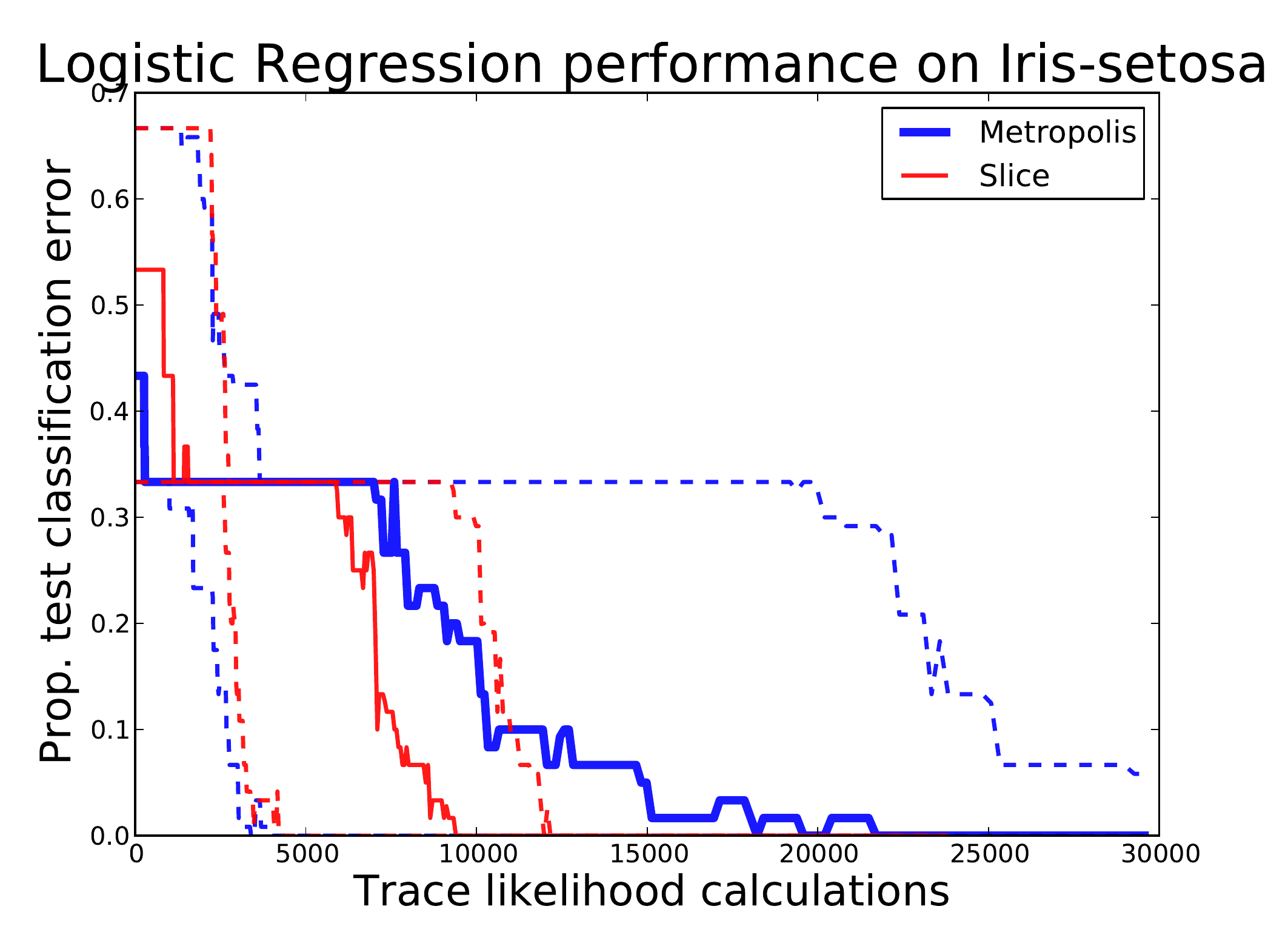}
        \end{minipage}
        ~ 
        \begin{minipage}[t]{0.32\textwidth}
                \centering
                \includegraphics[width=\textwidth]{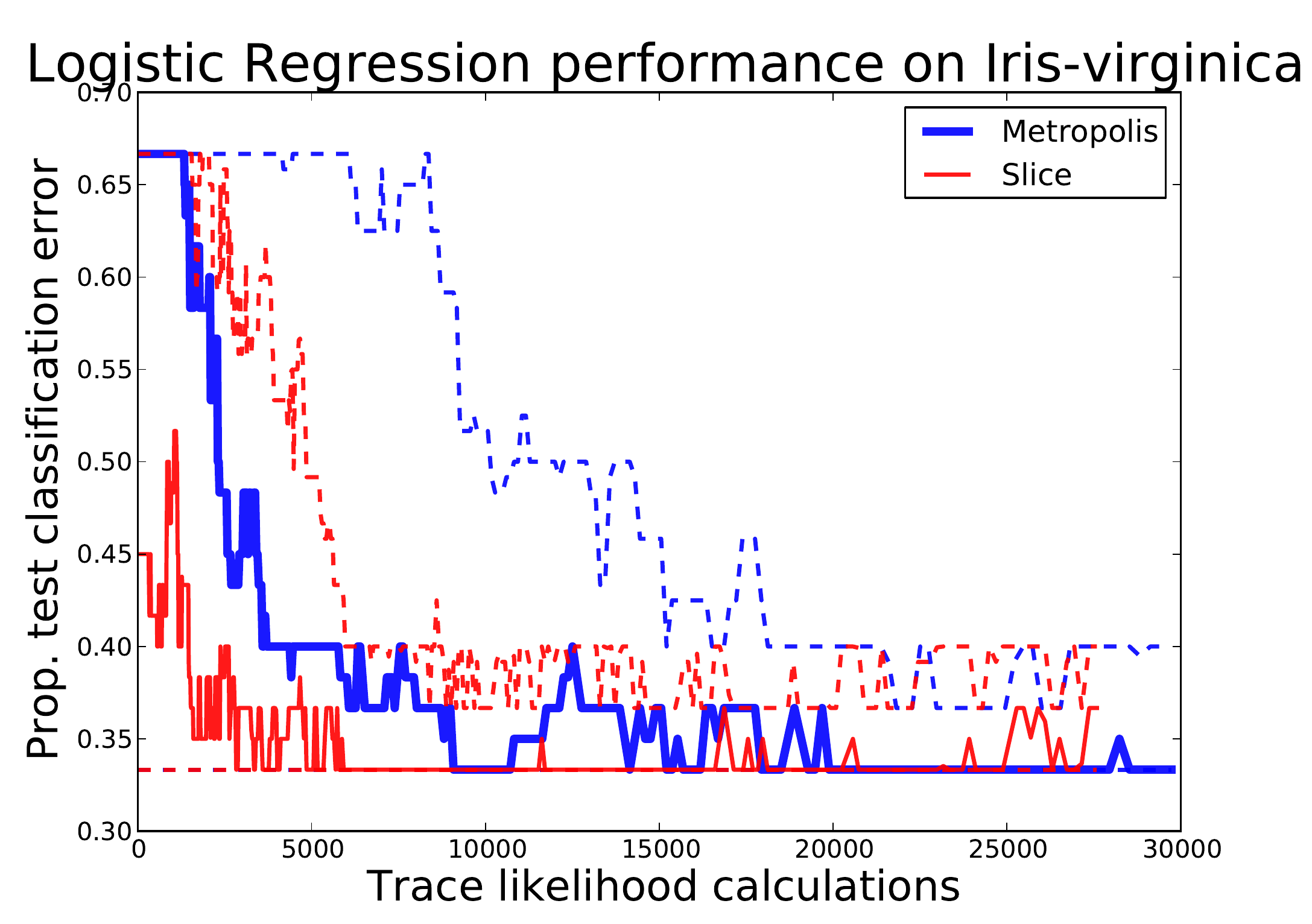}
        \end{minipage}
        ~
        \begin{minipage}[t]{0.32\textwidth}
                \centering
                \includegraphics[width=\textwidth]{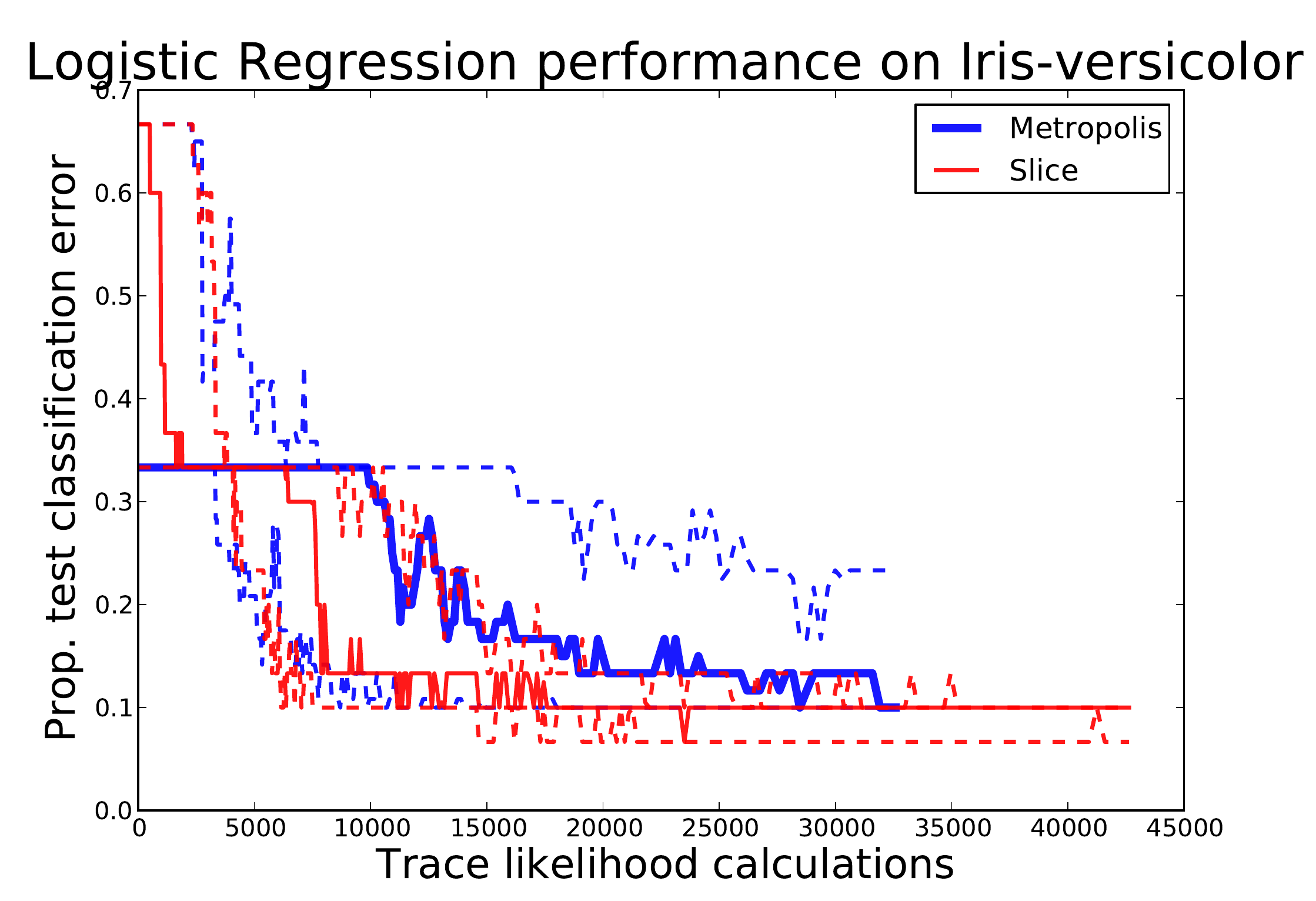}
        \end{minipage}
\caption{Median (solid), 25\% and 75\% quartiles (dashed) convergence rates of logistic regression on Iris dataset}
  \label{fig:logRegConv}
\end{figure*}

\begin{figure*}[!ht]
  \label{fig:nn}
        \centering
        \begin{minipage}[t]{0.32\textwidth}
                \centering
                \includegraphics[width=\textwidth]{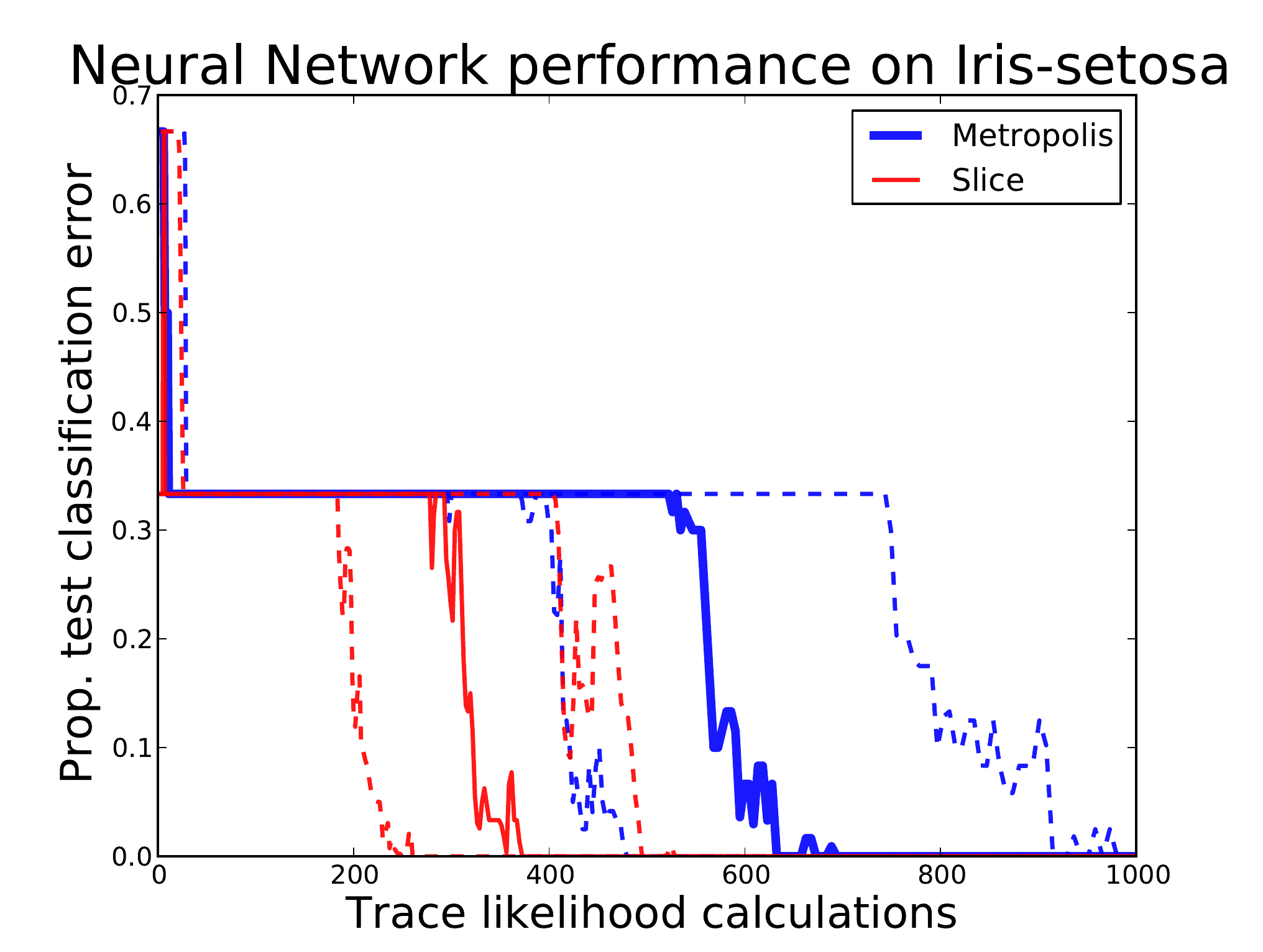}
        \end{minipage}
        ~ 
        \begin{minipage}[t]{0.32\textwidth}
                \centering
                \includegraphics[width=\textwidth]{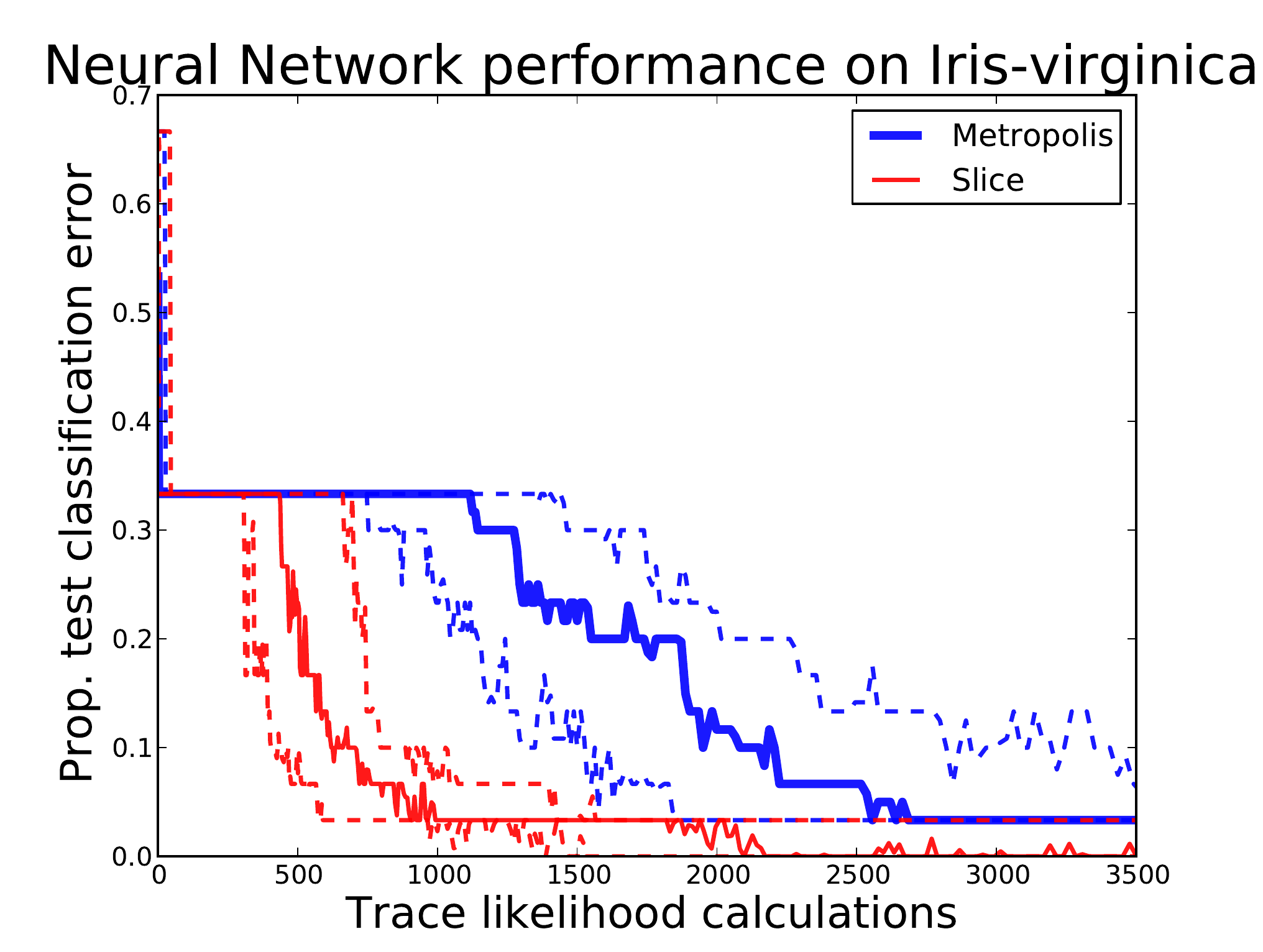}
        \end{minipage}
        ~
        \begin{minipage}[t]{0.32\textwidth}
                \centering
                \includegraphics[width=\textwidth]{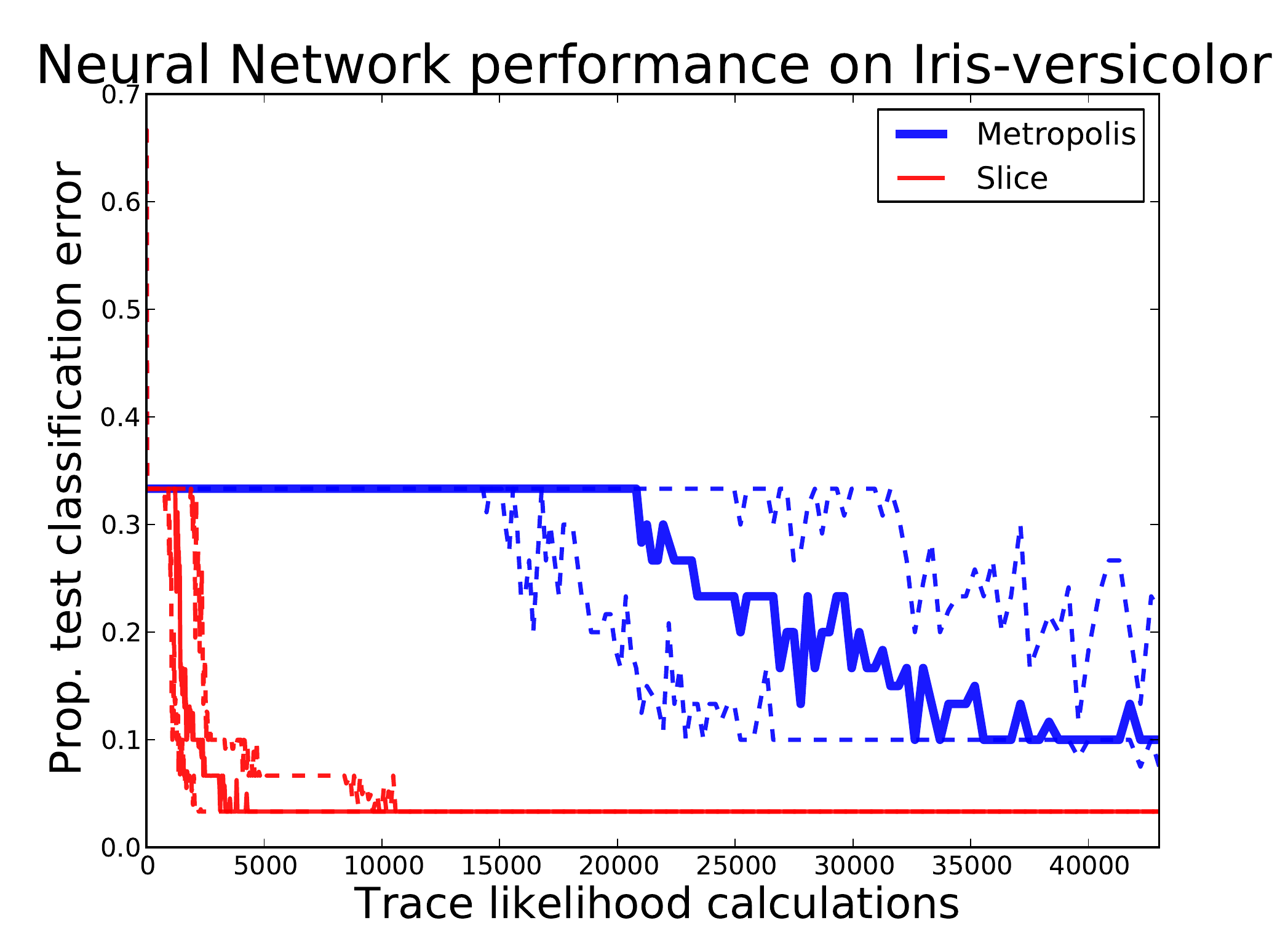}
        \end{minipage}
\caption{Median (solid), 25\% and 75\% quartiles(dashed) convergence rate of a Bayesian NN on the Iris dataset}
\label{fig:nnIrisConv}
\end{figure*}

First we perform logistic regression on the well known Iris dataset, obtained from \cite{Bache+Lichman:2013}. The result is shown in Figure \ref{fig:logRegConv} and shows that Slice sampling converges significantly faster than MH. Secondly we train a small Bayesian neural network on the same dataset. Our neural network is composed of an input layer of the same dimensionality as the data (4), two hidden layers of 4 and 2 neurons respectively and an output layer of a single neuron (since we are treating the task as a binary classification problem). In Figure \ref{fig:nnIrisConv} we see that slice sampling does significantly better than the MH baseline on this task. We also notice that the harder the inference problem is, the more the margin by which slice sampling outperforms grows. Iris-setosa, on which the performance gap is smallest, is linearly separable from the other 2 classes, which are not linearly separable from one another. In the case of Iris-versicolor, we were not able to run the engine long enough for Metropolis to match slice's performance, as Metropolis is still lagging behind even after running 10 times longer than slice.

\section{Related Work}

Our slice sampling inference engine is based on the slice sampling method presented by \cite{neal2003slice}, and influenced by the computer friendly slice sampler shown in \cite{mackay2003information}. 

Slice sampling techniques have been applied to a wide range of inference problems (\cite{neal2003slice}) and are used in some of the most popular PPLs, such as BUGS (\cite{lunn2009bugs}) and STAN (\cite{stan-manual:2014}). However, the slice samplers these languages employ are not exposed directly to the user, but instead only used internally by the language. Therefore, the slice samplers present in these languages are not intended to generalise to all models and, specifically, make no mention of trans-dimensionality corrections. \cite{poon2006sound}, also proposes a slice sampling based solution, this time to Markov Logic problems. However this algorithm is focused on deterministic or near-deterministic inference tasks and so bears little resemblance to our approach.

The most similar use-case to ours would be in Venture (\cite{mansinghka2014venture}). Here however slice sampling is only mentioned in passing, amongst other inference techniques the system could support. No details, nor discussions of trans-dimensionality, are present.
  
\section{Conclusion}

We have introduced and made available StocPy, the first general purpose Python Probabilistic Programming Language. We have shown the benefits StocPy offers both to users and designers of PPLs, namely flexibility, clarity, succintness and ease of prototyping. 

We have also implemented a novel, slice sampling, inference engine in StocPy. To our knowledge this is the first general purpose slice sampling inference engine and the first slice sampling procedure that solves the problem of trans-dimensionality. 

We have empirically evaluated this slice sampling engine and shown that the potential benefits far outweight the potential costs, when compared to single-site Metropolis. While Metropolis works well on very small models where the prior and the posterior are similar, slice provides substantial benefits as the distributions diverge. Additionally, on the models where Metropolis performs best slice only experiences a constant slowdown due to its overhead, whereas when Metropolis performs poorly the performance difference can be arbitrarily large. 

Finally, we have provided comparisons with Anglican which show promising results despite slice sampling not beeing optimised for runtime speed. A full benchmarking of the systems remains to be performed.

\clearpage
\bibliographystyle{abbrvnat}
\bibliography{slice}

\end{document}


%

%

\twocolumn[

\aistatstitle{Supplementary Material for ``Slice Sampling for Probabilistic Programming''}
\aistatsauthor{Razvan Ranca \And Zoubin Ghahramani}

\aistatsaddress{University of Cambridge \And University of Cambridge } ]

\section{Detailed slice-sampling pseudocode}

Here we present the detailed pseudocode for one possible slice sampling implementation. Algorithm \ref{alg:sliceSamp} is the high-level logic of slice sampling which was also presented in the paper. 

The other algorithms expand upon this, showing the lower level operations. Specifically, Algorithm \ref{alg:stepOut} shows how we could perform the operation described in line 5 of Algorithm \ref{alg:sliceSamp}, by exponentially increasing the $[x_l, x_r]$ interval. Similarly, Algorithm \ref{alg:sampNext} describes how the operation from line 6 of Algorithm \ref{alg:sliceSamp} is performed, namely by sampling a next $x$ placed on the already defined segment $[x_l, x_r]$ such that $x$ is choosen ``uniformly under the curve''.

\begin{algorithm}
  \caption{Slice Sampling}
  \begin{algorithmic}[1]
    \State $ \text{Pick initial } x_1 \text{ such that } P^*(x_1) > 0 $
    \State $ \text{Pick number of samples to extract } N $
    \For {$t = 1 \to N$} 
      \State $ \text{Sample a height } u \text{ uniformly from [0, } P^*(x_t)] $
      \State $ \text{Define a segment } [x_l,x_r] \text{ at height } u$
      \State $ \text{Pick }x_{t+1} \in [x_l, x_r] \text{ such that } P^*(x_{t+1}) > u$
   \EndFor
  \end{algorithmic}
  \label{alg:sliceSamp}
\end{algorithm}

\begin{algorithm}
  \caption{Return appropriate $[x_l, x_r]$ interval, given current sample $x$ and its height $u$}
  \begin{algorithmic}[1]
    \Function{StepOut}{$x, u$}
    \State $ \text{Pick small initial step width } w_i $
    \While {$P^*(x - w) < u \text{ or } P^*(x + w) < u$}
      \State $ w_i = w_i/2$
    \EndWhile
    \State $ w = w_i $
    \While {$P^*(x - w) > u$} 
      \State $ w = w * 2 $
    \EndWhile
    \State $ x_l = x - w $
    \State $ w = w_i $
    \While {$P^*(x + w) > u$} 
      \State $ w = w*2 $
    \EndWhile
    \State $ x_r = x + w $
    \State \Return {$x_l, x_r$}
  \EndFunction
  \end{algorithmic}
  \label{alg:stepOut}
\end{algorithm}

\begin{algorithm}
  \caption{Given current sample $x$ and its height $u$, get next sample $x_n$ from the interval $[x_l, x_r]$ such that $P^*(x_n) > u$}
  \begin{algorithmic}[1]
    \Function{SampNext}{$x, u, x_l, x_r$}
    \State $ \text{Sample $x_n$ uniformly from } [x_l, x_r] $
    \While {$ P^*(x_n) \leq u $}
      \If {$ x_n > x $}
        \State $ x_r = x_n $
      \Else
        \State $ x_l = x_n $
      \EndIf
      \State $ \text{Sample $x_n$ uniformly from } [x_l, x_r] $
    \EndWhile
    \State \Return {$x_n$}
  \EndFunction
  \end{algorithmic}
  \label{alg:sampNext}
\end{algorithm}

\section{True Posteriors}

\begin{figure*}[!ht]
  \centering
  \begin{minipage}[l]{1.0\columnwidth}
    \includegraphics[width=\textwidth]{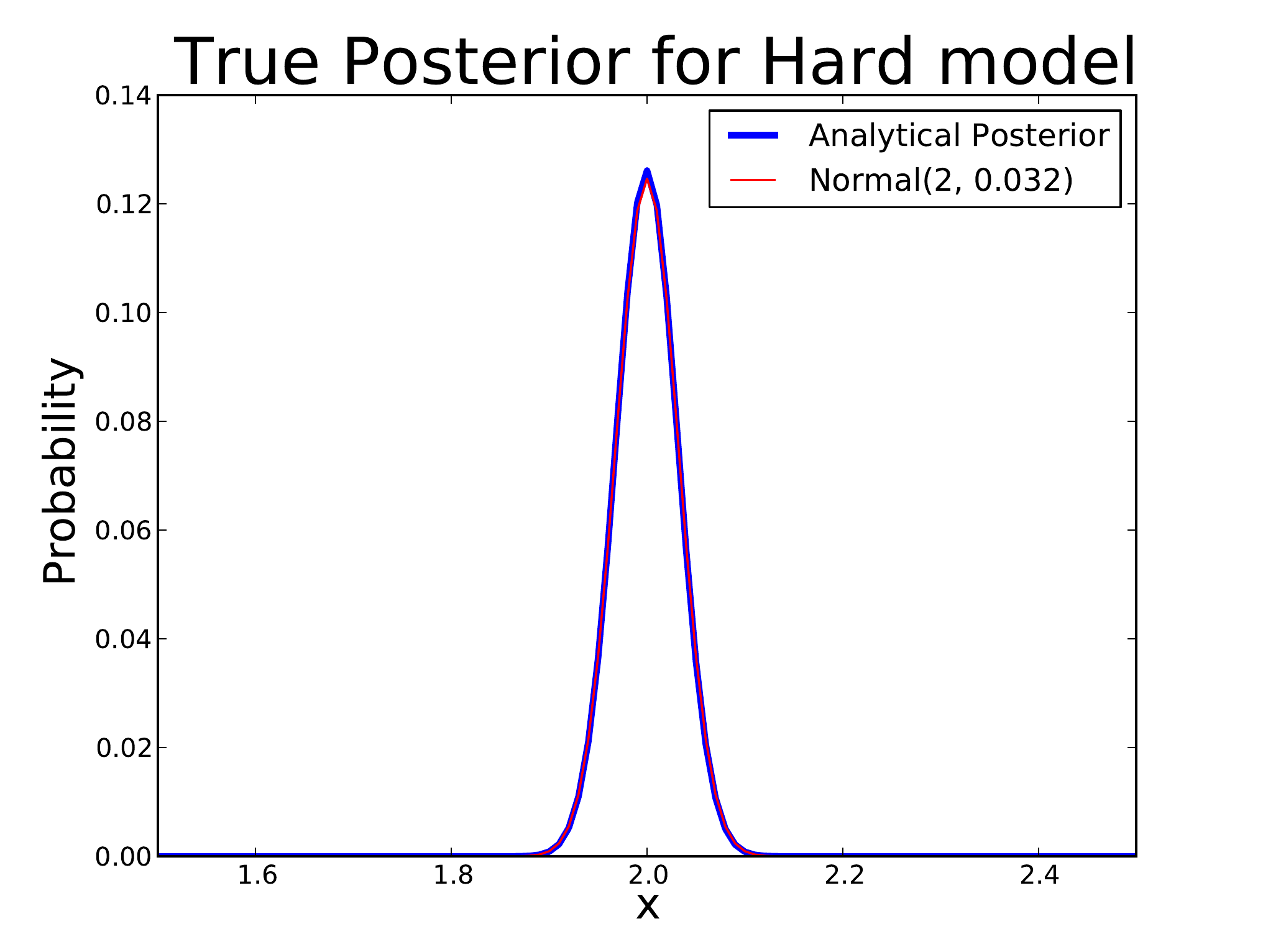}
    \caption{Analytically derived posterior for the Hard normal model, superimposed over a Gaussian}
    \label{fig:normHardPost}
  \end{minipage}
  ~
  \begin{minipage}[l]{1.0\columnwidth}
    \includegraphics[width=\textwidth]{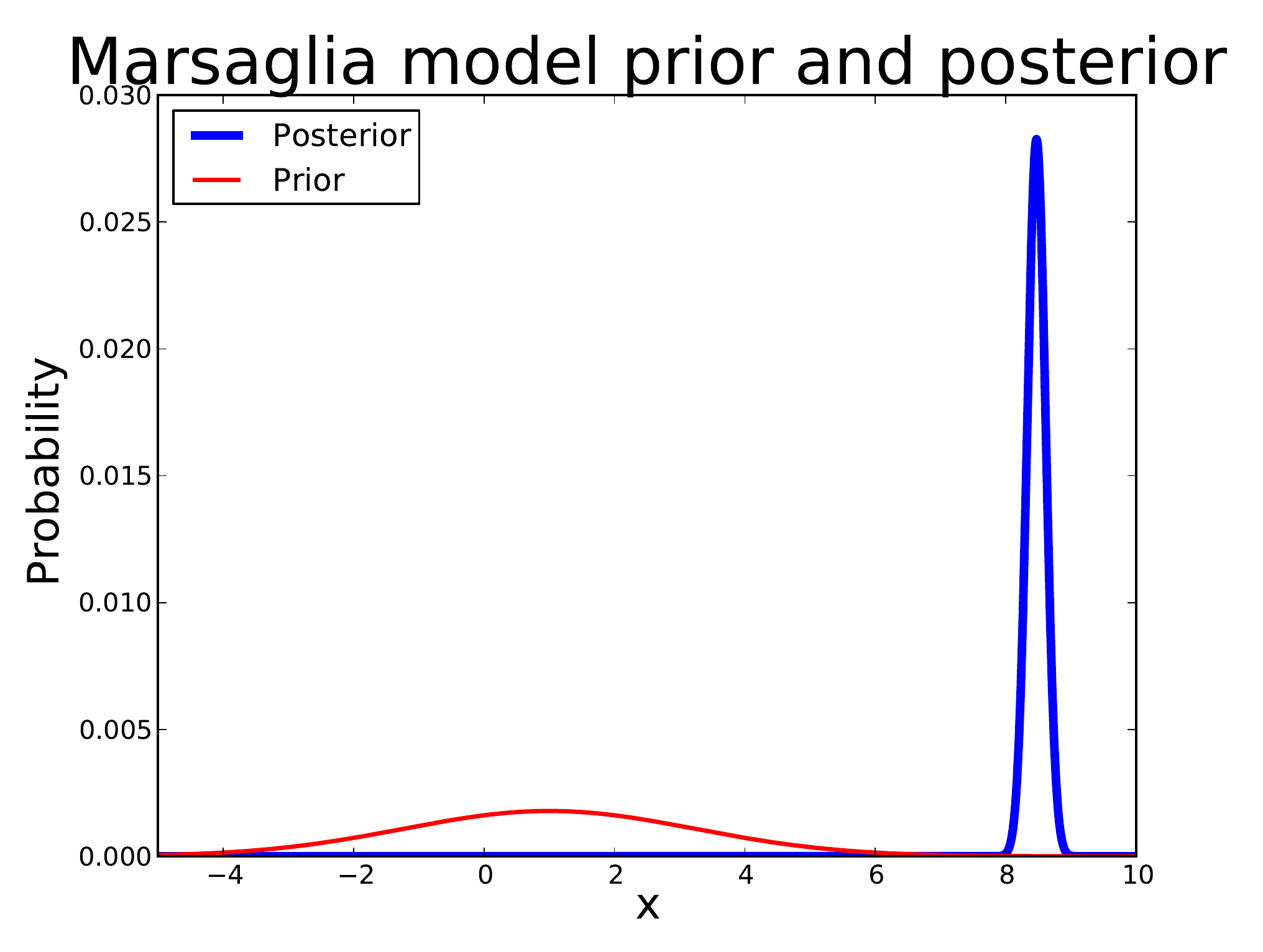}
    \caption{The posterior on the Marsaglia model is sufficiently different from the prior for slice sampling to outperform Metropolis}
    \label{fig:marsagliaPost}
  \end{minipage}
\end{figure*}

Here we present 2 true posteriors that support some claims made in the paper. In Figure \ref{fig:normHardPost}, we see that the ``Hard'' Gaussian mean inference problem results in an analytical posterior that is very close to a Gaussian with mean 2 and standard deviation 0.032. The second posterior, shown in Figure \ref{fig:marsagliaPost} presents the difference between the prior and posterior of the Marsaglia model. This figure gives an intuitive understanding of why Metropolis is outperformed by slice sampling on the Marsaglia model. It is easy to imagine how bad Metropolis will do in this model if it only samples from the prior.

\section{Additional Sample Programs in StocPy and Anglican}

For completeness and as to get a better idea of the ``feel'' of StocPy we show the remaining 2 Anglican models that were not presented in the paper. These models are a Dirichlet Process Mixture (Figure \ref{alg:dpMixModel}) and a Marsaglia (Figure \ref{alg:marsaglia}). We also show these models' implementation in Anglican, for comparison's sake.

\begin{figure*}[!ht]
  \centering
  \begin{minipage}[l]{1.0\columnwidth}
\begin{lstlisting}[language=Python, basicstyle=\ttfamily\scriptsize, label=alg:dpmPy]
def dpMix():
  crp = stocPy.CRP(1.72)
  sds, ms, cs = {}, {}, {}
  for ind, ob in stocPy.readCSV("dpm-data.csv"):
    c = crp(ind)
    cs[ind] = c
    if c not in ms:
      sds[c] = math.sqrt(10 * stocPy.stocPrim("invgamma", (1,0,10)))
      ms[c] = stocPy.stocPrim("normal", (0, sds[c]))
    stocPy.normal(ms[c], sds[c], cond=ob)
  stocPy.observe(cs, name="u")
  stocPy.observe(len(ms), name="K")

\end{lstlisting}
  \end{minipage}
  \hfill{}
   \begin{minipage}[r]{1.0\columnwidth}
\begin{lstlisting}[language=Lisp, basicstyle=\ttfamily\scriptsize, label=alg:dpmAng]
[assume class-generator (crp 1.72)]
[assume get-class (mem (lambda (n) (class-generator)))]
[assume get-var (mem (lambda (c) (/ 1.0 (gamma 1 10))))]
[assume get-std (lambda (c) (sqrt (get-var c)))]
[assume get-mean (mem (lambda (c) (normal 0 (sqrt (* 10 (get-var c))))))]
[assume u (lambda () (list (get-class 1) (get-class 2) (get-class 3) (get-class 4) (get-class 5) (get-class 6) (get-class 7) (get-class 8) (get-class 9) (get-class 10))) ]
[assume K (lambda () (count (unique (u))))]
[observe-csv "dpm-data.csv" (normal (get-mean (get-class $1)) (get-std (get-class $1))) $2)]
[predict (u)]
[predict (K)]
\end{lstlisting}
  \end{minipage}
  \caption{DP Mixture model expressed in StocPy (left) and Anglican (right)}
  \label{alg:dpMixModel}
\end{figure*}

\begin{figure*}[!ht]
  \centering
  \begin{minipage}[l]{1.0\columnwidth}
\begin{lstlisting}[language=Python, basicstyle=\ttfamily\scriptsize, label=alg:marsagliaPy]
def marsaglia(mean, std):
  x = stocPy.uniform(-1, 1)
  y = stocPy.iuniform(-1, 1)
  s = x*x + y*y
  if s < 1:
    return mean + (std * (x * math.sqrt(-2 * (math.log(s) / s))))
  return marsaglia(mean, std)

def marsagliaMean():
  mean = marsaglia(1, math.sqrt(5))
  std = math.sqrt(2)
  stocPy.normal(mean, std, cond=9)
  stocPy.normal(mean, std, cond=8)
  stocPy.observe(mean, name="mean")
\end{lstlisting}
  \end{minipage}
  \hfill{}
   \begin{minipage}[r]{1.0\columnwidth}
\begin{lstlisting}[language=Lisp, basicstyle=\ttfamily\scriptsize, label=alg:marsagliaAng]
[assume marsaglia-normal (lambda (mu std)
          (begin
            (define x (uniform-continuous -1.0 1.0))
            (define y (uniform-continuous -1.0 1.0))
            (define s (+ (* x x) (* y y))) 
            (if (< s 1) 
                (+ mu (* std (* x (sqrt (* -2.0 (/ (log s) s))))))
                (marsaglia-normal mu std))))]
[assume std (sqrt 2)]
[assume mu (marsaglia-normal 1 (sqrt 5))]
[observe (normal mu std) 9]
[observe (normal mu std) 8]
[predict mu]

\end{lstlisting}
  \end{minipage}
  \caption{Marsaglia model expressed in StocPy (left) and Anglican (right)}
  \label{alg:marsaglia}
\end{figure*}

As in the paper, we can see that on the more complex model StocPy manages to be more succint, largely thanks to the Python in-built operators.
Additionally, clarity is a subjective perception, but we would argue that the StocPy formulations are friendlier for people without a strong CS or technical background, such as domain experts.